\pdfoutput=1

\documentclass[10pt,twocolumn,letterpaper]{article}
\usepackage[pagenumbers]{cvpr} %

\usepackage[T1]{fontenc}
\usepackage{cuted}
\usepackage{currfile}
\usepackage{caption}
\usepackage{amsmath,amssymb,amsthm}
\usepackage{bm}
\usepackage{graphicx}
\usepackage{array}
\usepackage{tabularx}
\usepackage{multirow}
\usepackage{makecell}
\usepackage[table]{xcolor}
\usepackage{colortbl}
\usepackage{longtable}
\usepackage{arydshln}
\usepackage{pifont}
\usepackage{xspace}
\usepackage{placeins}
\usepackage{algorithm}
\usepackage{algpseudocode}
\usepackage{tikz}
\usepackage{refcount}

\definecolor{rowblue}{HTML}{E5EFFA}   %
\definecolor{defgray}{gray}{0.92}     %
\definecolor{grptext}{gray}{0.38}     %
\definecolor{dgreen}{HTML}{1A823F}
\definecolor{dred}{HTML}{BF2929}
\definecolor{citeblue}{HTML}{1A3E8C}  %
\definecolor{figcyan}{HTML}{006E7F}   %
\definecolor{tabpurple}{HTML}{5B2C83} %
\definecolor{eqorange}{HTML}{B05A00}  %
\definecolor{secgreen}{HTML}{1B6B2C}  %
\definecolor{appgreen}{HTML}{2E5B4C}  %
\definecolor{algred}{HTML}{9B1B1B}    %
\definecolor{thmbrown}{HTML}{6B4226}  %
\definecolor{lemteal}{HTML}{16877D}   %
\definecolor{propmag}{HTML}{8C1D57}   %
\definecolor{defnavy}{HTML}{1C2B5E}   %
\definecolor{urlrose}{HTML}{A3456A}   %

\theoremstyle{plain}
\newtheorem{proposition}{Proposition}
\newtheorem{corollary}{Corollary}
\newtheorem{lemma}{Lemma}
\theoremstyle{definition}

\DeclareMathOperator{\ED}{ED}
\DeclareMathOperator{\logit}{logit}
\DeclareMathOperator{\clip}{clip}
\newcommand{\E}{\mathbb{E}}
\newcommand{\R}{\mathbb{R}}
\newcommand{\ind}{\mathbf{1}}

\newcommand{\method}{SibPO\xspace}

\newcommand{\tablestyle}{\scriptsize\setlength{\tabcolsep}{1.2pt}\renewcommand{\arraystretch}{1.12}}
\newcolumntype{Y}{>{\centering\arraybackslash}X}
\newcommand{\grp}[2]{\multicolumn{#1}{l}{\textcolor{grptext}{\textit{#2}}}}
\newcommand{\hd}[1]{\textbf{#1}}
\newcommand{\up}{$\uparrow$}
\newcommand{\dn}{$\downarrow$}
\newcommand{\one}{${\to}1$}

\newcommand{\thickhline}{\noalign{\global\setlength{\arrayrulewidth}{0.9pt}}\hline\noalign{\global\setlength{\arrayrulewidth}{0.4pt}}}

\definecolor{cvprblue}{rgb}{0.21,0.49,0.74}
\usepackage[pagebackref,breaklinks,colorlinks,hypertexnames=false,citecolor=citeblue,linkcolor=secgreen,urlcolor=urlrose,filecolor=secgreen]{hyperref}
\makeatletter
\DeclareRobustCommand{\coloredref}[3]{{\hypersetup{linkcolor=#1,filecolor=#1}%
  \edef\@xrurl{\getrefbykeydefault{#2}{url}{}}\edef\@xranc{\getrefbykeydefault{#2}{anchor}{}}%
  \expandafter\hyper@@link\expandafter{\@xrurl}{\@xranc}{#3}}}
\makeatother
\newcommand{\figref}[1]{\coloredref{figcyan}{#1}{Figure~\ref*{#1}}}
\newcommand{\tabref}[1]{\coloredref{tabpurple}{#1}{Table~\ref*{#1}}}
\newcommand{\secref}[1]{\coloredref{secgreen}{#1}{Section~\ref*{#1}}}
\newcommand{\appref}[1]{\coloredref{appgreen}{#1}{Appendix~\ref*{#1}}}
\newcommand{\eqnref}[1]{\coloredref{eqorange}{#1}{Eq.~(\ref*{#1})}}
\newcommand{\propref}[1]{\coloredref{propmag}{#1}{Proposition~\ref*{#1}}}
\newcommand{\corref}[1]{\coloredref{thmbrown}{#1}{Corollary~\ref*{#1}}}
\newcommand{\lemref}[1]{\coloredref{lemteal}{#1}{Lemma~\ref*{#1}}}

\newcommand{\algoref}[1]{\coloredref{algred}{#1}{Algorithm~\ref*{#1}}}
\makeatletter
\newcommand{\xrrecord}[2]{\@ifundefined{r@#1}{\global\@namedef{r@#1}{#2}}{}}
\makeatother

\def\confName{CVPR}
\def\confYear{2027}

\title{Comparing Each Generation with Independent Siblings Gives Video World Model Post-Training a Proper Objective for Outcome Frequencies}

\author{%
Xin Li\\
University of Southern California
\and
Yunlong Jia\\
Meta
\and
Yiwen Chen\\
Shanghai University of Sport
}

\hypersetup{pdftitle={\thetitle},pdfauthor={Xin Li; Yunlong Jia; Yiwen Chen},
  pdfinfo={YunlongJiaAffiliation={Meta}, YunlongJiaEmail={yunlongj@usc.edu},
    YunlongJiaPosition={Researcher, AI Solution and Automation},
    YunlongJiaEducation={University of Southern California, Viterbi School of Engineering},
    YunlongJiaOpenReview={https://openreview.net/profile?id=\string~Yunlong\_Jia2},
    AuthorReviewStatus={Pending confirmation of the final author list and order by the other authors}}}
\begingroup
  \immediate\pdfobj stream attr {/Type /Metadata /Subtype /XML} file {author-metadata.xmp}
  \pdfcatalog{/Metadata \the\pdflastobj\space 0 R}
\endgroup

\begin{document}
\maketitle
\begin{abstract}
Video world models are used as simulators, so when a first frame leaves several physical outcomes open, repeated generations should produce each outcome as often as it occurs in the data.
Physics post-training scores every generation against the single recorded future of its history, which makes single videos more faithful, but the expected score is linear in the model distribution and is maximized by always producing the most likely outcome.
The log-likelihood of the recorded outcome is maximized by the data distribution, and its gradient weights agreement with the record by the inverse of how often the model produces the generated outcome, a frequency that other generations of the same history with independent noise, its siblings, can estimate.
To train with this weight from a finite group, we propose SibPO, which performs \textbf{Sib}ling-paired \textbf{P}olicy \textbf{O}ptimization: siblings estimate a truncated inverse-frequency weight for agreement with the record, and a second term subtracts the same weight times the agreement with the siblings, which makes the expected objective a strictly proper divergence over outcome cells at every group size.
On PisaBench, SibPO lowers the KS distance to simulated landing times by 9.1--10.2\% relative to the strongest method outside its reward family on three of four video models without degrading single videos, and it lowers the error of estimated robot policy success rates by 13\%.
\end{abstract}
\section{Introduction}
\label{sec:intro}

\makeatletter\global\@topnum=0\makeatother %
\begin{figure}[t]
  \centering
  \includegraphics[width=\linewidth]{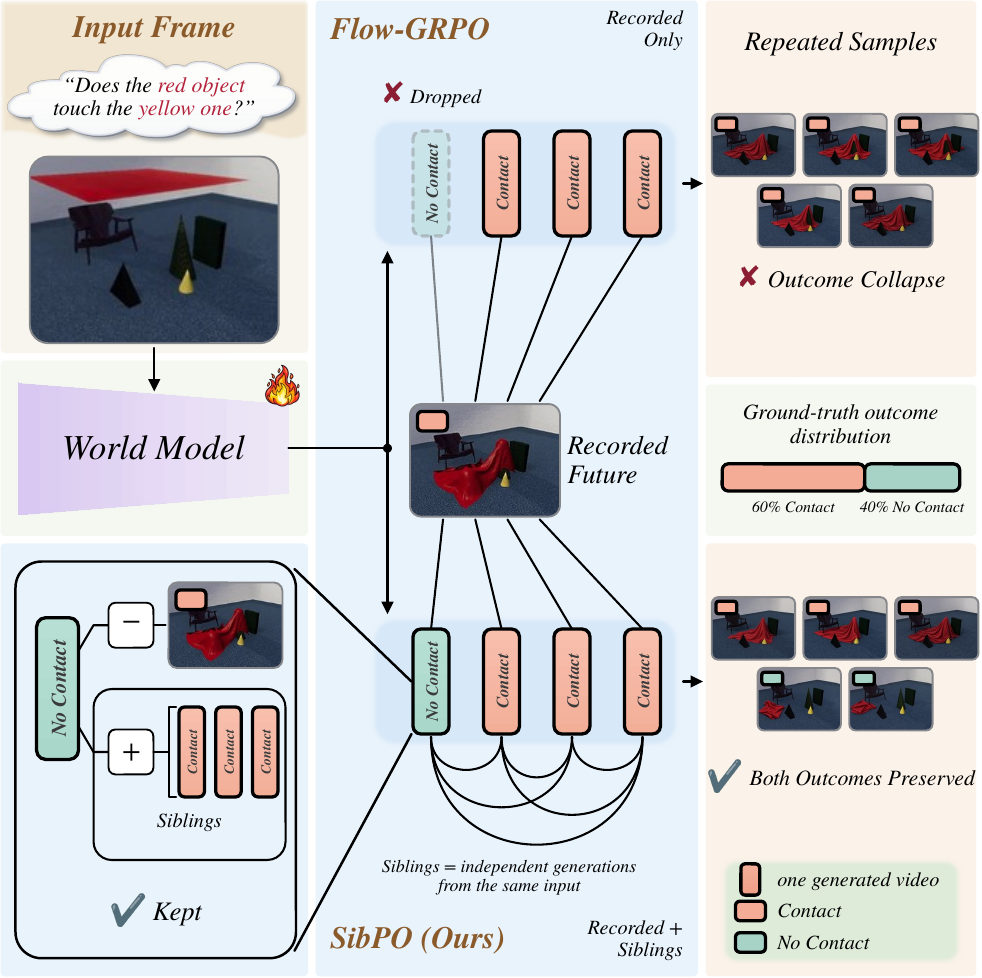}
  \caption{\textbf{What each generation is compared with decides the outcome frequencies after post-training.}
  The frame leaves open whether the red cloth will touch the yellow object, and $60\%$ of such futures show contact. Each block is the readout of one generated video, and five samples per method are only illustrative.
  Flow-GRPO compares each generation only with the recorded future, so the no-contact outcome loses on average and disappears.
  \method also compares each generation with its siblings, the other generations of the same input. Its reward, in the simplest form, subtracts the mismatch with the record ($-$) and adds the mismatch with the siblings ($+$), so the rare outcome is kept.}
  \label{fig:teaser}
\end{figure}

Video generation models continue a scene from a single frame with plausible motion, so they serve as world models that sample futures of an observed history, for instance to estimate robot policy success rates from imagined rollouts.
A single image, however, leaves part of the physics open. The frame in \figref{fig:teaser} does not show whether the falling red cloth will touch the yellow object, so the correct output is a distribution over both outcomes.
Evaluation has therefore moved from whether each video looks plausible~\cite{brooks2024sora,bruce2024genie,nvidia2025cosmos} to whether repeated generations reproduce the probabilities of the physically possible outcomes~\cite{pu2026pawbench,sadeghi2026calibench}.

Post-training makes video world models physically more faithful along two routes. The first route scores each generation against the recorded future of its history, through perceptual error~\cite{wu2025rlvrworld}, agreement of masks, optical flow or depth~\cite{li2025pisa} or preferences among recorded continuations~\cite{li2026dynamicsboost}, often with group policy gradients~\cite{liu2025flowgrpo,wang2026worldr1}, and makes single videos more faithful than likelihood fine-tuning.
The second route counters the resulting loss of diversity with diversity rewards~\cite{liu2026diversegrpo,kazimi2026dppgrpo} or a divergence to the initial model~\cite{li2026dphrl}, and the weight of this term decides how much spread survives.
Neither route gives the outcome frequencies a target that the data define.
No system evaluated on PAWBench reproduces the reference outcome probabilities~\cite{pu2026pawbench}, and the physics post-trained checkpoints of PISA~\cite{li2025pisa} lower the real-video trajectory error from $7.6$ to $6.7$--$7.5\times10^{-2}$ while their landing-time spread shrinks from $0.81$ to $0.63$--$0.74$ of the simulated width (\appref{app:released}).

\figref{fig:teaser} shows why the first route concentrates. If $60\%$ of valid continuations end in contact, a model that produces contact with probability $p$ agrees with a random record with probability $0.6p+0.4(1-p)=0.4+0.2p$, which grows with $p$. Every reward for agreeing with the record is therefore maximized by always producing contact, and a KL or diversity weight only limits how far the model moves toward this mode.

We therefore study the following problem. When a video world model is post-trained through an outcome readout, a frozen program that reads the outcome of interest from a video, and every history comes with a single recorded future, what determines the outcome frequencies of its repeated generations, and how can these frequencies reach those of the data while the single-video gains of readout rewards and the usual number of generations are kept?
The log-likelihood of the recorded outcome is maximized by the data distribution, and its gradient rewards agreement with the record in inverse proportion to how often the model already produces the generated outcome. Only other generations of the same history drawn with independent noise, which we call the siblings of a generation, can estimate this frequency, and GRPO-style post-training already draws them.

Two obstacles remain. A finite group estimates only a truncated inverse frequency, capped by the group size, and dividing the reward by it alone, as success-count normalization does~\cite{tajwar2026maxrl}, removes rare outcomes from the optimum in small groups.
Once the reward of a generation depends on its siblings, the advantage estimator, which turns the rewards of a group into the weights of the policy-gradient update, also becomes part of the objective. Group normalization or initial noise shared within a group~\cite{xue2025dancegrpo} then move the frequencies back toward the mode.

To make the data distribution the unique maximizer of the expected objective of every history at every group size, we propose \method, which trains each generation by \textbf{Sib}ling-paired \textbf{P}olicy \textbf{O}ptimization and changes only how GRPO-style post-training computes rewards and advantages.
Its reward has two terms. The first weights agreement with the record by the sibling estimate of the truncated inverse frequency. The second, which we call the sibling term, subtracts the same weight times the agreement of the generation with its siblings. It cancels the truncation bias, so the expected update follows a divergence between model and data outcome frequencies that vanishes only when they are equal, and a leave-one-out baseline, a batch-constant scale and independent sibling noise keep this objective intact.
Our contributions are as follows.
\begin{itemize}
  \setlength{\itemsep}{0pt}
  \item We show that, with one recorded future per history, what each generation is compared with sets the outcome frequencies of readout-based post-training, and how common advantage estimators move them toward the mode.
  \item We propose \method, whose sibling-estimated weight and sibling term make the data distribution the unique maximizer of the expected objective at every group size.
  \item On four video models and an autoregressive world model, \method has the lowest PisaBench KS distance on three and parity on the fourth without degrading single videos, interventions reproduce each predicted estimator effect, and policy success estimates improve by $13\%$.
\end{itemize}
\section{Related Work}
\label{sec:related}

Prior work asks how to make single generations follow physics, how to keep the diversity that reward optimization removes and, outside video generation, which target the outcome frequencies of an ambiguous input should have.

\paragraph{Physics post-training compares each generation with a recorded target, which improves single videos and leaves outcome frequencies unconstrained.}
Besides methods that inject physical attributes~\cite{wang2025wisa}, distill spatio-temporal relations~\cite{zhang2025videorepa}, optimize physical preferences~\cite{chen2025physhpo} or reward 3D and collision rules~\cite{wang2026worldr1,zhang2026physrvg}, three are closest to us: PISA~\cite{li2025pisa} rewards agreement of segmentation, optical flow or depth with recorded drops, RLVR-World~\cite{wu2025rlvrworld} runs GRPO with the L1 plus LPIPS distance to recorded frames, and DynamicsBoost~\cite{li2026dynamicsboost} prefers continuations conditioned on longer recorded prefixes.
Each reward scores one generation against one target, which fixes trajectory errors of single videos and drives an ambiguous history to its likeliest outcome (\secref{sec:analysis}). \method keeps their data and readouts and changes the comparison partner.

\paragraph{Reinforcement learning for generators made group policy gradients practical and counters diversity loss with weights that the data do not set.}
Flow-GRPO~\cite{liu2025flowgrpo} applied the group-relative advantage of GRPO~\cite{shao2024deepseekmath} to flow models through a stochastic sampler, DanceGRPO~\cite{xue2025dancegrpo} shares the initial noise within a group, and the resulting diversity loss~\cite{kirk2024rlhf} is countered with cluster-based~\cite{liu2026diversegrpo} or determinantal~\cite{kazimi2026dppgrpo} rewards, corrected reward directions~\cite{chen2026d2align} or a forward divergence to the initial policy~\cite{li2026dphrl}.
The closest analysis shows that group standard-deviation normalization makes the stated probabilities of language models overconfident, while RLOO~\cite{ahmadian2024rloo} and PPO~\cite{schulman2017ppo} stay calibrated~\cite{bereket2025uncalibrated}.
That analysis assumes independent rewards, while with sibling-coupled rewards even mean subtraction changes the objective (\secref{sec:theory}). A diversity weight sets a spread unrelated to the outcome distribution.

\paragraph{Proper scoring rules define the correct target, and their sample-based training uses quadratic or truncated forms.}
A strictly proper score such as the energy score~\cite{gneiting2007scoring} is maximized in expectation only by the true distribution and trains generative forecasters~\cite{pacchiardi2024scoring,shen2024engression,debortoli2025distributional,lang2024aifscrps} through differentiable samples.
Energy-based fine-tuning (EBFT)~\cite{jelassi2026ebft} matches mean features of language-model rollouts to one reference with a reward that adds alignment with the reference and subtracts similarity to the other rollouts, and MaxRL~\cite{tajwar2026maxrl} divides correctness rewards by the group success count to approach the log-likelihood of a fixed correct answer.
EBFT's quadratic weight does not grow as an outcome becomes rare, and MaxRL's truncated normalization becomes improper once the target outcome is random and removes rare cells in small groups. \method keeps the growing weight and removes the truncation bias with a sibling correction (\secref{sec:analysis}).
\section{Problem and Analysis}
\label{sec:problem}

\subsection{Problem Definition}
\label{sec:definition}

A video world model generates a future video from a history, an image or short clip with an optional action or prompt, and a simulation query asks for one outcome of this future, such as contact in \figref{fig:teaser}. Post-training sees each history once with one recorded future, and a solution succeeds if each generated outcome distribution approaches the true one at the usual number of generations without worse single videos.
Formally, a history $h\sim\mu$ comes with one recorded future $y\sim Q_h$.
The model generates $x=g_\theta(h,\xi)$, where $\xi$ collects all random numbers of the sampler, the initial noise and the noise of every step, and $P_h$ denotes the distribution of $x$.
An outcome readout $\varphi$ is a fixed program that maps a video to $\R^m$, in most cases to one of finitely many outcome cells such as contact and no contact, or the frame of first ground contact. A matching kernel $\kappa(a,b)\in[0,1]$ with $\kappa(a,a)=1$ states how strongly two readouts count as the same outcome. For outcome cells it is the indicator of equal cells.
With $P_h$ and $Q_h$ also denoting the induced distributions over outcomes, the goal is
\begin{equation}
  P_h = Q_h \quad \text{for $\mu$-almost every } h,
  \label{eq:goal}
\end{equation}
learned from pairs $(h,y)$.
Readouts are not differentiable, so updates use the score-function gradient $A\,\nabla_\theta\log\pi_\theta(\tau)$, with sampling trajectory $\tau$ of $x$ and an advantage $A$ computed from the rewards of the group.
In the running example, the outcome is binary with $q=Q_h(\text{contact})=0.6$ and $p=P_h(\text{contact})$.

\subsection{Analysis}
\label{sec:analysis}

\paragraph{A reward against the recorded future is linear in the model distribution.}
Any reward $s(x,y)$ that compares a generation with the record, such as the negative readout distance, has the expected value $\int P_h(dx)\,\E_{y}s(x,y)$.
It is linear in $P_h$ and hence maximized by a point mass at the mode of $Q_h$ for discrete outcomes and at a geometric median for continuous ones.
A KL penalty with weight $\beta$ stops at the reference tilted by $\exp(\E_y s/\beta)$, which still moves the contact frequency of the running example toward $1$.
Likelihood training on recorded videos targets whole videos, where the few pixels that decide the outcome, such as a delayed landing, contribute little, so these errors persist. A solution must therefore keep the readout, which acts on these errors, and make the objective proper, that is, maximized only at $P_h=Q_h$.

\paragraph{Independent siblings supply the missing signal.}
For an outcome in a finite set, write $p_a=P_h(a)$ and $q_a=Q_h(a)$.
The expected log-likelihood of the recorded outcome, $\sum_a q_a\log p_a$, is maximized only at $P_h=Q_h$. Its gradient $\sum_a q_a\nabla p_a/p_a$ is the expected agreement with the record, weighted by $1/p_a$, the inverse of the frequency with which the model produces the generated outcome, which siblings can estimate. The probability that $m$ independent siblings all miss outcome $a$ is $(1-p_a)^m$, and these probabilities sum over $m\geq0$ to $1/p_a$. The $G-1$ siblings in a group of $G$ estimate the terms with $m\leq G-1$ without bias. Keeping the terms up to an order $M$ gives the truncated inverse-frequency weight
\begin{equation}
  w_M(p)=\sum_{m=0}^{M}(1-p)^m=\frac{1-(1-p)^{M+1}}{p}.
  \label{eq:weight}
\end{equation}
It equals $1$ for $M=0$ and grows with $M$ up to $w_M(0)=M+1$ for an outcome that the model no longer produces.
Weighting agreement by $w_M$ alone, as success-count normalization does~\cite{tajwar2026maxrl}, equalizes $q_a w_M(p_a)$ across the kept outcomes and removes every outcome with $(M+1)q_a$ below that level, so rare outcomes vanish in small groups (\secref{sec:theory}).

\paragraph{The sibling term removes the truncation bias.}
Subtracting from the reward the same weight times the frequency $p_a$ with which siblings reproduce the generated outcome gives the expected update
\begin{equation}
  \textstyle\sum_a w_M(p_a)\,(q_a-\lambda p_a)\,\nabla p_a,\qquad \lambda=1,
  \label{eq:reward}
\end{equation}
where $\lambda$ is the weight of the subtracted sibling term. The weight now multiplies the gap $q_a-p_a$ between data and model, so the update stops exactly at $q_a=p_a$, however the weight is truncated.
\eqnref{eq:reward} is the gradient of $-D_M(Q_h\Vert P_h)$, the Bregman divergence of a separable convex function with curvature $w_M>0$, which vanishes only at $P_h=Q_h$ for every $M$. We call an objective with this property strictly proper.
At $M=0$ it equals the Brier score, the quadratic form of feature matching~\cite{jelassi2026ebft}, and as $M$ grows it approaches $\mathrm{KL}(Q_h\Vert P_h)$, so the pull on a nearly dropped outcome grows by up to $w_M(0)=M+1$.

\paragraph{Common estimator choices change the comparison.}
\eqnref{eq:reward} assumes independent siblings and an advantage that leaves the sibling terms unchanged. Published estimators break this in four ways: the group mean of GRPO and RLOO puts the current sample into its own baseline through the sibling terms of the other rewards, the group standard deviation reproduces the single-comparison advantage for binary outcomes, initial noise shared within a group~\cite{xue2025dancegrpo} lets siblings agree on what that noise decides, and partners from other histories constrain only pooled frequencies. Each pulls the stationary point toward the mode (\secref{sec:theory}), so a proper objective therefore requires independent siblings of the same history with $\lambda=1$, a baseline that excludes the current sample and an advantage scale that does not depend on the group.
All three concern only the reward and the advantage.
\begin{figure*}[t]
  \centering
  \includegraphics[width=\textwidth]{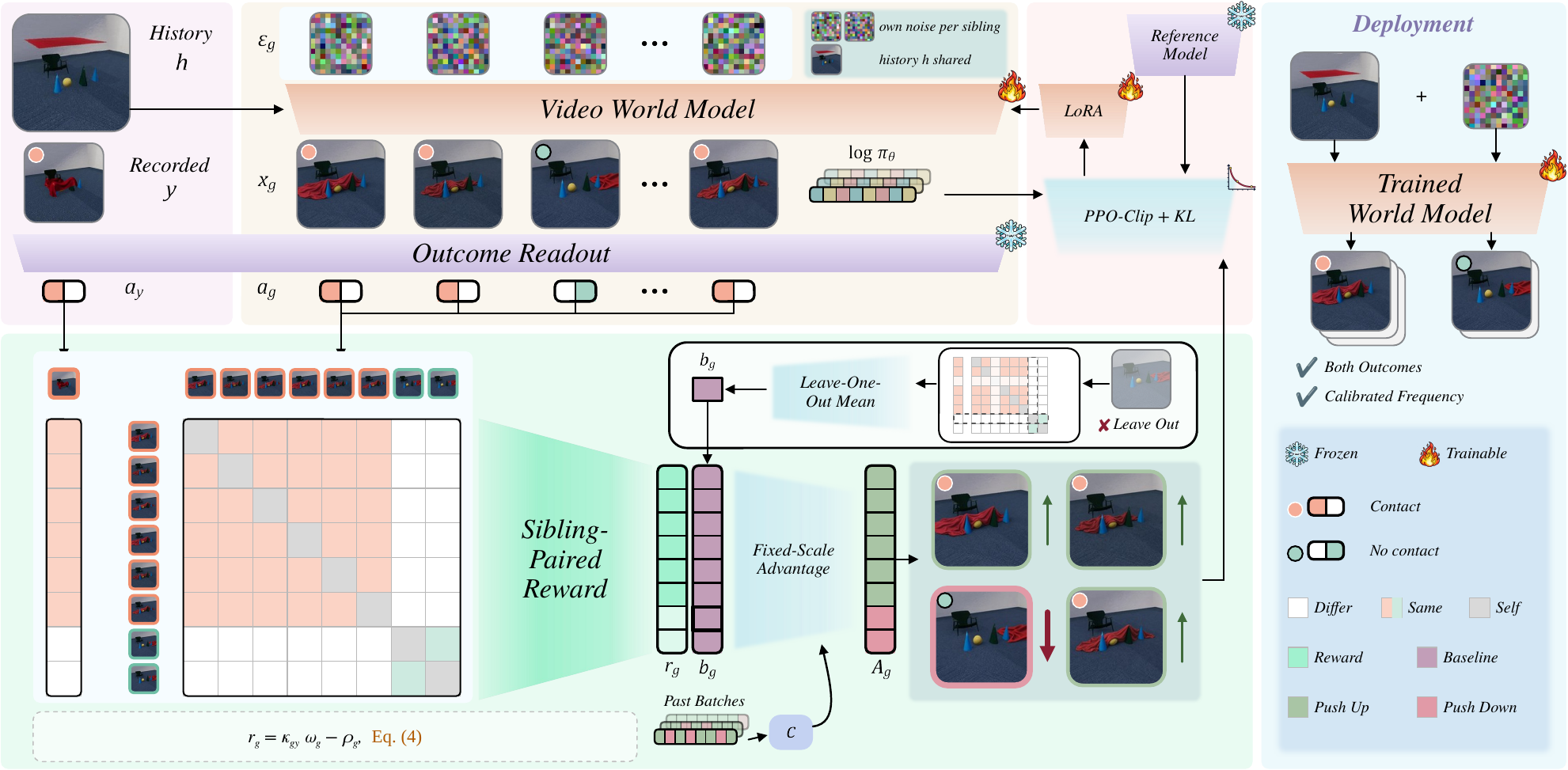}
  \caption{\textbf{Overview of \method on one training group.}
  All $G=8$ generations $x_g$ share the history $h$ and draw their own noise $\varepsilon_g$ (shared noise is only a comparison setting in \secref{sec:ablation}). A frozen readout maps each video to a two-cell outcome vector, contact (orange) or no contact (teal).
  The matrix pairs the eight generations, grouped by outcome, with each other and with the record (left column): a cell takes the outcome color when the two agree ($\kappa=1$) and stays white when they differ.
  The reward $r_g$ of \eqnref{eq:m1} weights agreement with the record by $\omega_g$, which grows with the white cells of row $g$, and subtracts the sibling term $\rho_g$. The baseline $b_g$ recomputes the rewards without $x_g$.}
  \label{fig:method}
\end{figure*}

\section{Method}
\label{sec:method}

\paragraph{Overview.}
\method turns the three requirements of \secref{sec:analysis} into an advantage computation for GRPO-style post-training and leaves the model, the data, the readout and the number of generations unchanged (\figref{fig:method}).
For each history the model generates $G$ siblings, a frozen readout extracts their outcomes and that of the record, the sibling-paired reward compares every generation with the record and its siblings to define the proper target (\secref{sec:reward}), and a leave-one-out baseline with a fixed scale (\secref{sec:baseline}) and independent sibling noise (\secref{sec:procedure}) keep the expected update on this target.

\subsection{Sibling-Paired Reward}
\label{sec:reward}

The reward estimates both factors of \eqnref{eq:reward}, the weight $w_M$ at the model frequency of the generated outcome and this weight times the frequency, from the other generations of the group.
For a history $h$ with recorded future $y$ and siblings $x_1,\dots,x_G$, write $a_g=\varphi(x_g)$ and $a_y=\varphi(y)$ for the readouts, $\kappa_{gj}=\kappa(a_g,a_j)$ for the agreement of generations $g$ and $j$ and $\kappa_{gy}=\kappa(a_g,a_y)$ for the agreement of $g$ with the record.
The product of the miss indicators $1-\kappa_{gj}$ over $m$ siblings estimates $(1-p)^m$ without bias, and its average over all $m$-subsets is the elementary symmetric polynomial $e_m$ of these indicators divided by the number of subsets. Summing up to order $M$ estimates $w_M$, and the reward of sibling $g$ is
\begin{equation}
\begin{aligned}
  r_g &= \kappa_{gy}\,\omega_g-\lambda\,\rho_g,\qquad
  \omega_g=\sum_{m=0}^{M}\frac{e_m(\{1-\kappa_{gj}\}_{j\neq g})}{\binom{G-1}{m}},\\
  \rho_g &= \frac{1}{G-1}\sum_{j\neq g}\kappa_{gj}\sum_{m=0}^{M}\frac{e_m(\{1-\kappa_{gk}\}_{k\neq g,j})}{\binom{G-2}{m}},
\end{aligned}
  \label{eq:m1}
\end{equation}
with $\lambda=1$ and $M=G-2$, the largest order that the $G-2$ siblings left in $\rho_g$ support.
Given $a_g$, $\omega_g$ estimates $w_M$ at the model frequency of $a_g$ without bias, and the sibling term $\rho_g$ estimates this frequency times the same weight by pairing one agreeing sibling $j$ with a weight estimated from the others.
The first term pays more for matching the record with a rare outcome, and the second removes what a frequent outcome already collects.
In the group of \figref{fig:method}, six of eight generations and the record show contact. For a contact generation, two of its seven siblings differ, which gives $\omega_g=1+\tfrac27+\tfrac1{21}=1.33$, $\rho_g=1.00$ and $r_g=+0.33$. A no-contact generation has six differing siblings and $\omega_g=4.00$, but it disagrees with the record and receives $r_g=-1.00$. With a no-contact record it would receive $4.00-1.00=+3.00$, so averaged over records with contact probability $0.6$ it gains $+0.60$ and a contact generation $-0.20$, which raises the rare outcome toward $0.4$.
For $M=0$, $\omega_g=1$ and $r_g$ is the reward of quadratic feature matching~\cite{jelassi2026ebft}, an energy score~\cite{gneiting2007scoring} for $\kappa=1-d/d_{\max}$.

The readout and kernel determine the target and are fixed before training. Every video readout here is a finite outcome cell with the indicator kernel: continuous quantities such as landing positions fall into eight cells bounded by octiles of recorded training values, and an extra cell holds unreadable videos. Decoded robot futures use a Gaussian kernel with a bandwidth fixed before training (\appref{app:setup}). The training readout always differs from the evaluation judges.

\subsection{Leave-One-Out Baseline and Fixed Scale}
\label{sec:baseline}

A baseline reduces the variance of the update, but the group mean contains the current sample through the sibling terms of the other rewards.
\method therefore evaluates \eqnref{eq:m1} on the group with $x_g$ removed, with $M$ reduced by one, and averages the resulting rewards,
\begin{equation}
  b_g = \frac{1}{G-1}\sum_{j\neq g} r_j^{(-g)},\qquad A_g = (r_g-b_g)/c.
  \label{eq:m2}
\end{equation}
It depends only on the other siblings and the record, so it removes shared variation without bias. In the example group, $b_g=0.00$ for a contact generation and $0.14$ for a no-contact one, which gives the advantages $+0.33$ and $-1.14$ shown in \figref{fig:advantage} for $c=1$.
The scale $c$ is a running average of $|r_g-b_g|$ over previous batches, so it is constant within a batch, does not reweight outcomes inside a group and gives every history equal weight.

\subsection{Independent Siblings and Training Procedure}
\label{sec:procedure}

The sibling terms estimate model frequencies only for siblings that are independent given the history. All siblings of a group share the history, encoded once per group, and each draws its own initial noise and its own noise at every sampling step. Autoregressive models share the key-value cache of the history and sample all future tokens independently.
A training step generates $G\geq3$ siblings for each of $P$ histories with a stochastic sampler as in Flow-GRPO~\cite{liu2025flowgrpo}, stores the log-probability of every sampling transition or token, computes advantages by \eqnref{eq:m1} and \eqnref{eq:m2} and updates LoRA~\cite{hu2022lora} parameters with the PPO clipped objective~\cite{schulman2017ppo} and a KL penalty to the reference model (\algoref{alg:sibpo}).
The cost equals that of GRPO, and deployment uses the training sampler (\appref{app:deploy}).
\section{Theory}
\label{sec:theory}

We make \secref{sec:analysis} exact for one history. The results concern the expected update before the KL penalty and PPO clipping, for outcomes in finite cells with the indicator kernel, under three assumptions: (A1)~given $h$, the recorded future is independent of all model randomness, (A2)~$M\leq G-2$, and (A3)~trajectory log-probabilities are differentiable with interchangeable expectation and gradient, as for Gaussian sampler steps and tokens.
Proofs are in \appref{app:proofs}.

\begin{proposition}[Pairing determines the stationary point]
\label{prop:pairing}
Fix $h$, let $W_M'=w_M$ and $V_M'(p)=p\,w_M(p)$, and let the baseline not depend on the current generation. Under (A1)--(A3), the expected update of \eqnref{eq:m1} is the gradient of an objective $F$ that depends on how the siblings are drawn.
\begin{enumerate}
\setlength{\itemsep}{0pt}
\item If the siblings are independent generations of $h$,
\begin{equation}
F=\textstyle\sum_a\big[q_a W_M(p_a)-\lambda V_M(p_a)\big],
\label{eq:F1}
\end{equation}
which for $\lambda=1$ equals $-D_M(Q_h\Vert P_h)+\mathrm{const}$.
\item If the siblings share random numbers $u$ that do not depend on $\theta$, such as the initial noise, $F$ is the average of \eqnref{eq:F1} over the noise-conditional distributions $P_{h,u}$, and for $\lambda=1$
\begin{equation}
F=-D_M(Q_h\Vert P_h)-\Gamma^M_u+\mathrm{const},
\label{eq:F2}
\end{equation}
with $\Gamma^M_u=\E_u D_M(Q_h\Vert P_{h,u})-D_M(Q_h\Vert P_h)\geq 0$.
\item If the siblings come from other histories with a fixed pooled distribution, or if $M=0$ and $\lambda=0$, the objective is linear in $P_h$.
\end{enumerate}
Among these pairings, only independent siblings with $\lambda=1$ make $P_h=Q_h$ the unique maximizer for every $M$.
\end{proposition}

The reward formula alone thus does not fix the target: the same reward is proper with independent siblings, gains a penalty toward outcomes that the shared noise decides when siblings share noise, and becomes linear with partners from other histories.

For the success-count member ($\lambda=0$, $M\geq1$), the maximizer satisfies $q_aw_M(p_a)=\mu$ for a common level $\mu$ and sets $p_a=0$ wherever $(M+1)q_a<\mu$. This bias is small for a binary outcome at $G=8$, but a six-frame landing distribution loses both tail frames at $G=4$ (\appref{app:proofs}). For a binary outcome and $M=0$, the stationary frequency is
\begin{equation}
p^\star=\clip_{[0,1]}\!\big(\tfrac12+(q-\tfrac12)/\lambda\big),
\label{eq:pstar}
\end{equation}
which moves from the mode at $\lambda=0$ to $q$ at $\lambda=1$, while for $M=G-2$, $\lambda\in\{0.5,1.5\}$ gives only $0.656$ and $0.645$ (\tabref{tab:family}).

\begin{figure*}[t]
  \centering
  \includegraphics[width=\textwidth,trim=0 104 0 0,clip]{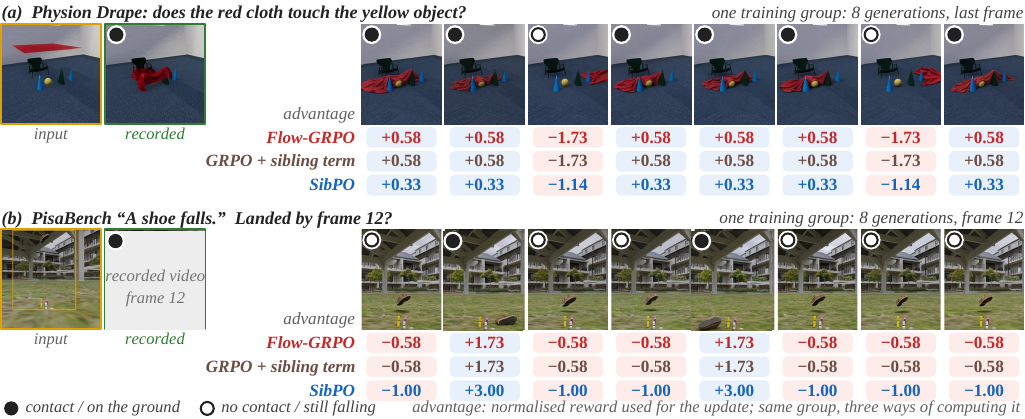}
  \caption{\textbf{Group standard-deviation normalization erases the sibling term, and the leave-one-out advantage of \method keeps it.}
  Advantages of three estimators for one Physion group of eight generations whose record shows contact (filled dots: contact). Flow-GRPO compares each generation only with the record. GRPO + sibling term adds the sibling term $-\rho_g$ of \eqnref{eq:m1}, which lowers the reward of a generation the more often its siblings share its outcome, but group mean and standard-deviation normalization returns the Flow-GRPO advantages (Part~3 of \propref{prop:group}). \method uses the leave-one-out baseline of \eqnref{eq:m2}.
  Averaged over records with contact probability $0.6$, the no-contact generation receives $-0.35$ under both normalized estimators and $+0.51$ under \method, which raises its frequency toward $0.4$.}
  \label{fig:advantage}
\end{figure*}

\begin{proposition}[Group advantages change the sibling terms]
\label{prop:group}
Let $x_1,\dots,x_G$ be independent generations of $h$ with rewards from \eqnref{eq:m1}, and let the update be $\frac1G\sum_g A_g\nabla_\theta\log\pi_\theta(\tau_g)$.
\begin{enumerate}
\setlength{\itemsep}{0pt}
\item If $A_g=(r_g-b_g)/c$ with $b_g$ a function of $(h,y,x_{-g})$ and $c>0$ independent of the group, as in \method, the expected update is $c^{-1}\nabla F$ with $F$ from \eqnref{eq:F1}.
\item With the mean of the other rewards (RLOO) or the group mean (GRPO) as baseline, the update for $M=0$ is, up to a positive factor, that of \eqnref{eq:F1} with $\lambda_{\mathrm{eff}}=\frac{G-2}{G-1}$.
\item For a binary outcome, $M\leq G-2$ and $\lambda\leq1$, group mean and standard-deviation normalization yields exactly the single-comparison advantages in every group.
\end{enumerate}
\end{proposition}

Part~1 justifies \secref{sec:baseline}, and Parts~2 and~3 show that standard normalizations change the objective.
Under Part~3, every group receives an expected logit update with the sign of $2q-1$, so without KL regularization $p\to1$ for every $G$ (\figref{fig:advantage}). For a kernel fixed before training, $P_h=Q_h$ remains a stationary point for every $M$ (\appref{app:proofs}).
\section{Experiments}
\label{sec:experiments}

\subsection{Experimental Setup}
\label{sec:setup}

\begin{table*}[t]
  \centering
  \caption{\textbf{Comparison with post-training methods on Cosmos-Predict2.5-2B~\cite{nvidia2025cosmospredict25}.}
  Same base weights, data, 150 updates and generation count. Held-out columns retrain without the falling family (Physion: all-but-one). $^\ddagger$Our split. $^\dagger$Spread-matched deployment. L2 and IoU $\times10^{-2}$, TVD $\times100$. ORO: PISA-ORO. Pear.: correlation with humans. Dyn.: VBench-I2V dynamic degree. GPU-h: A100 training hours. Bold: best per column.}
  \label{tab:main}
  \tablestyle
  \begin{tabularx}{\textwidth}{l||YY|YYYY|Y||Y|Y|Y|Y||YY|Y|Y||Y}
    \thickhline
    \multirow{3}{*}{\hd{Method}} & \multicolumn{7}{c||}{\hd{Trained mechanisms}} & \multicolumn{4}{c||}{\hd{Held-out mechanisms}} & \multicolumn{4}{c||}{\hd{Single-video quality}} & \multirow{3}{*}{\hd{GPU-h}} \\
    \cline{2-16}
     & \multicolumn{2}{c|}{PisaBench} & \multicolumn{4}{c|}{Physion} & Kubric$^\ddagger$ & Pisa. & Physion & PAW. & Cali. & \multicolumn{2}{c|}{PisaBench} & Phys.-IQ & VBench & \\
    \cline{2-16}
     & KS\dn & IQR\one & Brier\dn & Pear.\up & $\kappa$\up & Acc.\up & KS\dn & KS\dn & Brier\dn & TVD\dn & mnTV\dn & L2\dn & IoU\up & Score\up & Dyn.\up & \\
    \hline\hline
Base model & 0.585 & 0.71 & 0.255 & 0.352 & 0.189 & 60.2 & 0.412 & 0.585 & 0.255 & 30.5 & 0.606 & 12.0 & 9.3 & 25.3 & 40.7 & 0.0 \\
\cdashline{1-17}
SFT & 0.302 & 0.81 & 0.250 & 0.375 & \textbf{0.201} & 60.9 & 0.301 & 0.503 & 0.253 & 29.4 & 0.628 & 9.4 & 11.0 & 25.4 & 42.0 & 21.6 \\
\hline\grp{17}{Rewards against the recorded future}\\
RLVR-World~\cite{wu2025rlvrworld} & 0.331 & 0.56 & 0.276 & 0.361 & 0.185 & 59.9 & 0.322 & 0.524 & 0.272 & 31.9 & 0.641 & 9.0 & 11.5 & 24.3 & 39.3 & 21.1 \\
Flow-GRPO~\cite{liu2025flowgrpo} & 0.329 & 0.55 & 0.281 & 0.349 & 0.178 & 59.6 & 0.326 & 0.529 & 0.276 & 31.6 & 0.648 & 8.9 & 11.7 & 25.8 & 40.1 & 21.5 \\
ORO (Seg)~\cite{li2025pisa} & 0.318 & 0.60 & 0.274 & 0.344 & 0.177 & 59.8 & 0.316 & 0.517 & 0.270 & 31.8 & 0.637 & 9.0 & \textbf{11.9} & 24.1 & 40.6 & 31.5 \\
ORO (Flow)~\cite{li2025pisa} & 0.326 & 0.57 & 0.277 & 0.350 & 0.176 & 59.5 & 0.321 & 0.522 & 0.273 & 31.2 & 0.643 & 8.7 & 11.1 & 26.1 & 43.2 & 29.8 \\
ORO (Depth)~\cite{li2025pisa} & 0.321 & 0.58 & 0.272 & 0.357 & 0.181 & 59.9 & 0.318 & 0.520 & 0.269 & 30.9 & 0.645 & \textbf{8.5} & 11.0 & 26.3 & 42.4 & 30.6 \\
DynamicsBoost~\cite{li2026dynamicsboost} & 0.287 & 0.70 & 0.258 & 0.366 & 0.193 & 60.6 & 0.297 & 0.497 & 0.259 & 30.1 & 0.624 & 9.0 & 11.5 & \textbf{27.2} & \textbf{45.4} & 24.8 \\
\hline\grp{17}{Diversity-preserving post-training}\\
DiverseGRPO~\cite{liu2026diversegrpo} & 0.322 & 0.58 & 0.270 & 0.364 & 0.184 & 59.8 & 0.320 & 0.521 & 0.267 & 29.2 & 0.619 & 9.0 & 11.3 & 23.5 & 43.6 & 21.9 \\
DPH-RL~\cite{li2026dphrl} & 0.276 & 0.73 & 0.252 & 0.368 & 0.192 & \textbf{61.3} & 0.292 & 0.489 & 0.256 & 29.6 & 0.621 & 9.1 & 11.2 & 23.0 & 41.6 & 29.1 \\
\hline\grp{17}{Controls for spread and for the inverse-frequency weight}\\
Base model$^\dagger$ & 0.571 & \textbf{0.96} & 0.251 & 0.348 & 0.183 & 59.8 & 0.398 & 0.571 & 0.251 & \textbf{29.0} & \textbf{0.594} & 12.6 & 8.8 & 24.6 & 44.9 & 0.0 \\
Flow-GRPO$^\dagger$ & 0.297 & 0.95 & 0.262 & 0.346 & 0.176 & 59.3 & 0.305 & 0.507 & 0.264 & 30.4 & 0.631 & 9.9 & 10.6 & 25.0 & 44.0 & 21.5 \\
\method, $M=0$ & 0.256 & 0.80 & 0.243 & 0.377 & 0.196 & 61.1 & 0.281 & 0.471 & 0.249 & 29.3 & 0.603 & 9.0 & 11.4 & 25.8 & 41.8 & 21.6 \\
\hline\hline
\rowcolor{rowblue}\method (ours) & \textbf{0.241} & 0.86 & \textbf{0.239} & \textbf{0.386} & 0.198 & 61.0 & \textbf{0.274} & \textbf{0.463} & \textbf{0.246} & 29.1 & 0.598 & 8.9 & 11.5 & 25.4 & 42.1 & 21.6 \\
    \thickhline
  \end{tabularx}
\end{table*}

\begin{figure*}[t]
  \centering
  \includegraphics[width=\textwidth]{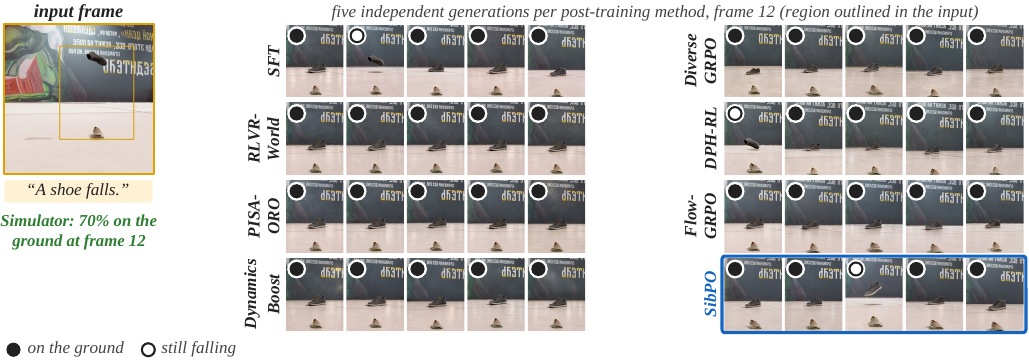}
  \caption{\textbf{Rewards against the recorded future collapse an ambiguous landing, and \method keeps its minority outcome.}
  Five generations per method on Cosmos-Predict2.5-2B at frame 12 of one PisaBench input, where the simulator places $70\%$ of objects on the ground.}
  \label{fig:qual-all}
\end{figure*}

\paragraph{Benchmarks and metrics.}
The PisaBench distribution protocol~\cite{li2025pisa} gives 50 frames of a falling object whose depth the image leaves open, each with an analytic landing-time distribution. From 128 generations we report the mean per-frame KS distance and the median ratio of generated to simulated interquartile range (IQR ratio), which isolates spread and should be one.
On 1,200 Physion~\cite{bear2021physion} test videos, eight continuations each give the fair Brier score, the only proper Physion metric, besides accuracy and human agreement.
Our Kubric split re-simulates 150 held-out scenes, PAWBench~\cite{pu2026pawbench} and CaliBench~\cite{sadeghi2026calibench} score untrained mechanisms such as dice, and single-video quality uses the PisaBench error on 361 real videos, Physics-IQ~\cite{motamed2025physicsiq} and VBench-I2V~\cite{huang2025vbenchpp}.

\paragraph{Training and compared methods.}
Post-training uses PISA falling objects, the eight Physion contact scenarios and Kubric~\cite{greff2022kubric} throws, bounces and collisions with one recorded future per history. For held-out mechanisms, every method is retrained without the falling family, and Physion follows the all-but-one protocol~\cite{bear2021physion}.
The main model is Cosmos-Predict2.5-2B~\cite{nvidia2025cosmospredict25}, and Wan2.2-TI2V-5B~\cite{wan2025wan}, CogVideoX1.5-5B-I2V~\cite{yang2025cogvideox}, Cosmos-Predict2.5-14B and iVideoGPT~\cite{wu2024ivideogpt} test generality. Compared methods cover rewards against the record (RLVR-World, Flow-GRPO, three PISA-ORO rewards, DynamicsBoost), diversity preservation (DiverseGRPO, DPH-RL) and flow-matching fine-tuning on recorded futures (SFT).
Two baselines isolate the weight: EBFT~\cite{jelassi2026ebft} pairs siblings through a quadratic score on frozen generator features, and \method with $M=0$ applies this score to the readout and differs from \method only in the weight.
Spread-matched controls ($^\dagger$) tune guidance and sampler noise to the simulated dispersion. All methods share base weights, data, LoRA rank, $G=8$, 150 updates and the stochastic sampler, and use three seeds with 95\% two-level bootstrap intervals. Training readouts use CoTracker3~\cite{karaev2025cotracker3} and SAM~2~\cite{ravi2025sam2} (\appref{app:setup}).

\subsection{Main Results}
\label{sec:main}

On the trained mechanisms of \tabref{tab:main}, \method reaches a PisaBench KS distance of $0.241$ against $0.276$ for DPH-RL, the strongest published method (difference $-0.035$, 95\% CI: $-0.052$--$-0.018$).
All post-training methods remove most of the landing delay of the base model, rewards against the record keep $0.55$--$0.70$ of the simulated IQR against $0.86$ for \method, and SFT keeps a wide spread ($0.81$) at the highest real-video error among post-training methods ($9.4\times10^{-2}$).
On binary Physion, rewards against the record raise the fair Brier score from $0.255$ to $0.258$--$0.281$, while \method lowers it to $0.239$, $0.011$ below SFT (95\% CI: $-0.017$--$-0.005$), and the less ambiguous Kubric split repeats the ordering ($0.274$ against $0.292$).

The control rows of \tabref{tab:main} attribute these gains.
The quadratic member ($M=0$) already reaches a KS distance of $0.256$ and a Brier score of $0.243$, below every method outside the \method family, so pairing with a proper score gives most of the gain and the weight adds $0.015$ and $0.004$.
Spread-matched controls reach the simulated IQR ($0.96$ and $0.95$), yet Flow-GRPO$^\dagger$ stays at a KS distance of $0.297$ with real-video error rising from $8.9$ to $9.9\times10^{-2}$.
Uniform widening misplaces the spread: it over-widens the narrowest quintile of frames (IQR ratio $1.36$) and leaves the widest too narrow ($0.69$), while the KS advantage of \method over Flow-GRPO grows with the simulated width (\appref{app:more}).

In the held-out columns of \tabref{tab:main}, \method reaches a PisaBench KS distance of $0.463$ against $0.489$ for DPH-RL (95\% CI: $-0.044$--$-0.008$), and on CaliBench Flow-GRPO carries its concentration over ($0.648$ against $0.606$ for the base model), while \method stays at the base-model level ($0.598$).

In \figref{fig:qual-all}, every method produces plausible single videos, but only SFT, DPH-RL and \method keep one of five generations in the air.

\subsection{Pairing Analysis}
\label{sec:ablation}

\begin{table}[t]
  \centering
  \caption{\textbf{Members of the sibling-paired reward family on Cosmos-Predict2.5-2B.}
  Theory: binary outcome, $q=0.65$, $G=8$, stationary frequency $p^\star$ and restoring update (Rest.) at $p=0.95$ relative to $M=0$, $\lambda=1$. EBFT has no outcome-level $p^\star$ (n/a).}
  \label{tab:family}
  \tablestyle
  \begin{tabularx}{\linewidth}{l||cc|YY|YYY}
    \thickhline
    \multirow{2}{*}{\hd{Reward}} & \multirow{2}{*}{\hd{$M$}} & \multirow{2}{*}{\hd{$\lambda$}} & \multicolumn{2}{c|}{\hd{Theory}} & \multicolumn{3}{c}{\hd{PisaBench / Physion}} \\
    \cline{4-8}
     & & & $p^\star$ & Rest.\up & KS\dn & IQR\one & Pear.\up \\
    \hline\hline
Single comparison & 0 & 0 & 1 & $-0.50$ & 0.334 & 0.54 & 0.352 \\
EBFT features~\cite{jelassi2026ebft} & 0 & 1 & n/a & n/a & 0.268 & 0.76 & 0.371 \\
Success count~\cite{tajwar2026maxrl} & 7 & 0 & 0.658 & 2.79 & 0.249 & 0.81 & 0.379 \\
\hline
\method, $M=0$ & 0 & 1 & 0.650 & 1.00 & 0.256 & 0.80 & 0.377 \\
\method, group mean & 6 & 1 & 0.651 & 3.52 & 0.243 & 0.85 & 0.384 \\
\method, $\lambda=0.5$ & 6 & 0.5 & 0.656 & 2.96 & 0.245 & 0.84 & 0.383 \\
\method, $\lambda=1.5$ & 6 & 1.5 & 0.645 & 4.12 & 0.242 & \textbf{0.87} & 0.385 \\
\rowcolor{defgray}\method & 6 & 1 & 0.650 & 3.54 & \textbf{0.241} & 0.86 & \textbf{0.386} \\
    \thickhline
  \end{tabularx}
\end{table}

In \tabref{tab:family}, the IQR ratio rises from $0.54$ for single comparison through $0.76$ for EBFT to $0.86$ for \method, and $\lambda\in\{0.5,1.5\}$ or a group-mean baseline change the KS distance by at most $0.004$, as $p^\star$ predicts. The weight lowers the KS distance by $0.015$ against $M=0$ (95\% CI: $-0.026$--$-0.004$), and at $G=8$ the sibling term lowers it by only $0.008$ against success counts, consistent with the small binary bias.

In \tabref{tab:group}, at 32 generations per update, success counts rise from a KS distance of $0.249$ at $G=8$ to $0.287$ at $G=4$ and narrow to an IQR ratio of $0.70$ as they drop tail cells, while \method changes by at most $0.007$, as \propref{prop:pairing} predicts, and leads at $G=4$ by $0.041$ (95\% CI: $0.025$--$0.057$).
Restarting from the collapsed Flow-GRPO checkpoint, the IQR ratio after 50 updates reaches $0.79$ for \method, $0.74$ for success counts and $0.67$ for $M=0$, the ordering of their restoring updates, while Flow-GRPO narrows to $0.53$.

Among the interventions of \tabref{tab:pairing}, the mean-of-others baseline stays within $0.003$ of \method, as exact enumeration for $M=G-2$ predicts (\appref{app:proofs}), and group standard-deviation normalization moves most of the way to Flow-GRPO ($0.293$--$0.322$ against $0.329$), completely only for two outcome cells (Part~3 of \propref{prop:group}).
Shared initial noise lowers the IQR ratio to $0.76$ and leaves the KS distance $0.021$ above \method, the direction of $\Gamma^M_u$ in \eqnref{eq:F2} (\appref{app:gamma}).

\figref{fig:crosshist} explains the strongest failure, partners from other histories, which lower the IQR ratio to $0.41$ yet reach the highest Physion accuracy ($61.2$). Both pairings match the pooled human rate of contact, but with cross-history partners each input collapses to its likelier outcome, because this pairing constrains only the pooled frequency.
Flipping $30\%$ of training readouts widens Flow-GRPO (IQR ratio $0.55$ to $0.66$) and narrows \method ($0.86$ to $0.79$), as \corref{cor:judge} predicts.

\subsection{Generality and Policy Evaluation}
\label{sec:scope}

\begin{figure}[t]
  \centering
  \includegraphics[width=0.1419\linewidth,trim=85.5 11.1 366.35 0,clip]{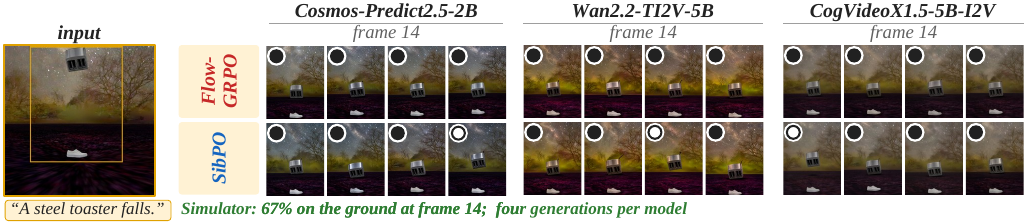}%
  \includegraphics[width=0.8581\linewidth,trim=250.5 11.1 0 0,clip]{figs/E6.pdf}
  \caption{\textbf{The pairing effect in two further model families.} Four generations at frame 14 for a falling toaster, which lands by this frame in $67\%$ of simulations.}
  \label{fig:families}
\end{figure}

In \tabref{tab:scope}, \method lowers the KS distance on Wan2.2 and at 14B to $0.281$ and $0.254$, against $0.309$ and $0.283$ for EBFT, the strongest method outside the \method family ($-9.1\%$ and $-10.2\%$, and $-10.1\%$ on Cosmos-2B). The weight contributes $0.020$ and $0.022$ on these narrower base models (IQR ratio $0.64$ and $0.59$ against $0.71$). On CogVideoX, whose base model is least concentrated ($0.75$), \method matches DPH-RL ($0.262$ against $0.271$, 95\% CI: $-0.024$--$0.006$).
In \figref{fig:families}, Flow-GRPO lands the toaster in all eight generations, while \method keeps one of four per model in the air, closer to the simulated $33\%$.

For robot policy evaluation, iVideoGPT follows the RT-1~\cite{brohan2023rt1} Real2Sim protocol of RLVR-World with five policies and ten SIMPLER tasks~\cite{li2024simpler}.
In \tabref{tab:ar}, \method lowers the error of estimated success rates to $11.2\times10^{-2}$ against $12.9\times10^{-2}$ for DPH-RL, the strongest published method (95\% CI: $-3.2$--$-0.2\times10^{-2}$), a reduction of $13\%$.
It keeps $94\%$ of the base dispersion against $68\%$ for RLVR-World, at a higher single-sample MSE ($0.520$ against $0.486\times10^{-2}$).
\section{Conclusion}
\label{sec:conclusion}

We have shown that, with one recorded future per history, what each generation is compared with determines the outcome frequencies after video world model post-training, and that coupled rewards make the advantage estimator part of the objective.
\method uses siblings to estimate the inverse-frequency weight of the outcome likelihood and to correct its truncation bias, which makes the data distribution the unique maximizer at every group size.
Its quadratic member already outperforms published post-training, and the weight restores dropped outcomes faster. The analysis covers any sampler with tractable log-probabilities trained from single observations through finite readout cells.

\clearpage
\setcounter{figure}{0}\renewcommand{\thefigure}{S\arabic{figure}}
\setcounter{table}{0}\renewcommand{\thetable}{S\arabic{table}}
\setcounter{equation}{0}\renewcommand{\theequation}{S\arabic{equation}}
\setcounter{lemma}{0}\renewcommand{\thelemma}{S\arabic{lemma}}
\setcounter{corollary}{0}\renewcommand{\thecorollary}{S\arabic{corollary}}
\setcounter{algorithm}{0}\renewcommand{\thealgorithm}{S\arabic{algorithm}}
\maketitlesupplementary
\appendix
\renewcommand{\dblfloatpagefraction}{0.85}
\renewcommand{\floatpagefraction}{0.85}
\renewcommand{\dbltopfraction}{0.75}

This supplementary material completes the main paper, whose sections, equations, propositions, figures and tables it cites by their main-paper numbers, while its own figures, tables and equations carry the prefix S.
\appref{app:proofs} proves the results of \secref{sec:theory}, \appref{app:setup} and \appref{app:bench} give the training procedure and the evaluation protocols, and \appref{app:ar} reports the full results on the further model families, at 14B scale and for policy evaluation.
\appref{app:released}, \appref{app:more}, \appref{app:gamma} and \appref{app:judge} expand the evidence for the mechanism with the released checkpoints, the pairing interventions, the measurement of shared-noise dispersion and the judge-accuracy sweep, and \appref{app:deploy} and \appref{app:monitor} cover deployment choices and training diagnostics.

\section{Proofs}
\label{app:proofs}

Throughout, $\pi_\theta(\tau)$ is the density of the sampling trajectory $\tau$ that produces a generation $X$ from history $h$, and all expectations condition on $h$.
Under (A3), any function $f$ of the generation that does not depend on $\theta$ satisfies the score-function identity~\cite{williams1992reinforce}
\begin{equation}
\E\big[f(X)\,\nabla_\theta\log\pi_\theta(\tau)\big]=\nabla_\theta\,\E\,f(X).
\label{eq:score}
\end{equation}
We write $F$ for the objective whose gradient equals the expected update, and constants are terms that do not depend on $\theta$.

\subsection{Unbiasedness of the Sibling Estimates}

\begin{lemma}
\label{lem:unbiased}
Let $\pi(a)=\E\,\kappa(a,\varphi(X))$ for an independent generation $X$. Given $a_g$, $\E\,\omega_g=w_M(\pi(a_g))$ for $M\leq G-1$ and $\E\,\rho_g=\pi(a_g)\,w_M(\pi(a_g))$ for $M\leq G-2$.
\end{lemma}
\begin{proof}
For a set $T$ of $m$ other siblings, independence gives $\E\prod_{j\in T}(1-\kappa_{gj})=(1-\pi(a_g))^m$, and $e_m/\binom{G-1}{m}$ averages these products over all $m$-subsets. In $\rho_g$, the factor $\kappa_{gj}$ uses a sibling outside the $m$-subset, so its expectation $\pi(a_g)$ multiplies $(1-\pi(a_g))^m$. Summing over $m\leq M$ gives \eqnref{eq:weight}.
\end{proof}

\subsection{Proof of Proposition~\ref*{prop:pairing}}

For finite outcomes and indicator $\kappa$, $\pi(a)=p_a$. By \lemref{lem:unbiased} and (A1), the expected reward of a generation with outcome $a$ is $w_M(p_a)(q_a-\lambda p_a)$, and \eqnref{eq:score} applied with the siblings held fixed gives the update $\sum_a w_M(p_a)(q_a-\lambda p_a)\nabla p_a$, which is the gradient of \eqnref{eq:F1}.
For $\lambda=1$ the derivative of $q_aW_M(p_a)-V_M(p_a)$ is $w_M(p_a)(q_a-p_a)=-\partial_{p_a}D_M(q_a\Vert p_a)$ for the Bregman divergence of the convex function $\phi_M$ with $\phi_M''=w_M>0$, so $F=-D_M(Q_h\Vert P_h)+\mathrm{const}$, which vanishes only at $P_h=Q_h$.
$M=0$ gives $w_0=1$ and the Brier score, and $w_M(p)\to1/p$ gives $\phi(p)=p\log p$ and $\mathrm{KL}(Q_h\Vert P_h)$.

\paragraph{Shared randomness.}
With shared $u$, the siblings are independent given $u$, so the argument applies to $P_{h,u}$ and the update averages \eqnref{eq:F1} over $u$, with gradients only through the transitions after $u$.
For $\lambda=1$ this gives \eqnref{eq:F2}.
$D_M(q\Vert p)$ is convex in $p$, since its second derivative $w_M(p)-w_M'(p)(q-p)$ is at least $w_M(p)+p\,w_M'(p)=(M+1)(1-p)^M>0$, so Jensen's inequality gives $\Gamma^M_u\geq0$.

\paragraph{Other histories and the linear cases.}
If siblings come from a fixed pooled distribution $\bar P$, their estimates do not depend on $P_h$, and the expected reward is linear in $P_h$. For $M=0$ and $\lambda=0$ the reward is the agreement with the record.
With $\lambda=0$ and $M\geq1$, $F=\sum_aq_aW_M(p_a)$ is strictly concave on the simplex since $W_M''=w_M'<0$, and its maximizer satisfies the Karush--Kuhn--Tucker conditions $q_aw_M(p_a)=\mu$ for $p_a>0$ and $q_aw_M(0)\leq\mu$ for $p_a=0$.
Because $w_M(0)=M+1$, every cell with $(M+1)q_a<\mu$ receives $p_a=0$, and the condition differs from $p=q$ for finite $M$.
For a binary outcome with $q=0.65$ this gives $p^\star=0.713$, $0.658$ and $0.650$ at $G=4$, $8$ and $16$ with $M=G-1$.
For the landing distribution $q=(0.05,0.15,0.30,0.25,0.15,0.10)$ over six consecutive frames, the maximizer is $(0,0.052,0.509,0.387,0.052,0)$ at $G=4$, $(0,0.139,0.390,0.316,0.139,0.017)$ at $G=8$ and $(0,0.155,0.332,0.275,0.155,0.083)$ at $G=16$, with KS distances to $q$ of $0.198$, $0.094$ and $0.050$, against $0.500$ for the mode.
Since $\bar P$ moves with the parameters, the cross-history result characterizes fixed points of the update.

\subsection{Proof of Proposition~\ref*{prop:group}}

Part~1 follows because $b_g$ depends on $(h,y,x_{-g})$, which is independent of $\tau_g$ given $h$, so $\E[b_g\nabla\log\pi_\theta(\tau_g)]=0$, and $c$ does not depend on the group.
Part~2 for $M=0$ is proved in \appref{app:energy}. For $M=G-2$ the other rewards depend on $x_g$ through their weights as well, and the quoted stationary points follow from exact enumeration of all group compositions.
For Part~3, rewards of \eqnref{eq:m1} depend on a binary group only through the outcome counts, so all generations with the same outcome receive the same reward, and the standardized advantages depend only on the counts and the sign of $r(y)-r(1-y)$.
With $n$ siblings sharing the generation's outcome among $G-1$, term $m$ of $\rho_g$ is $n/(G-1-m)$ times term $m$ of $\omega_g$, and $n\leq G-1-m$ whenever that term is nonzero, so $\rho_g\leq\omega_g$ with strict inequality when both outcomes occur.
Hence $r(y)\geq\omega-\lambda\rho>0\geq-\lambda\rho'=r(1-y)$ for $\lambda\leq1$, which is the sign of single comparison.

\subsection{Judge Accuracy under KL Regularization}

\begin{corollary}[Judge accuracy under KL regularization]
\label{cor:judge}
Consider a binary outcome, a judge that flips the outcome of a generated video with probability $\varepsilon_M<\tfrac12$ and that of a recorded video with probability $\varepsilon_D<\tfrac12$, and a KL penalty $\beta\,\mathrm{KL}(P_h\Vert P_{\mathrm{ref}})$.
Under single comparison, the optimum satisfies
\begin{equation}
\logit p^\star=\logit p_{\mathrm{ref}}+(1-2\varepsilon_M)(1-2\varepsilon_D)(2q-1)/\beta .
\label{eq:judge1}
\end{equation}
Under independent sibling pairing with $\lambda=1$ and any $M$, with $\tilde p=\varepsilon_M+(1-2\varepsilon_M)p$ and $\tilde q=\varepsilon_D+(1-2\varepsilon_D)q$, it satisfies
\begin{equation}
\logit p^\star=\logit p_{\mathrm{ref}}+\tfrac{1-2\varepsilon_M}{\beta}\big[w_M(\tilde p^\star)+w_M(1-\tilde p^\star)\big](\tilde q-\tilde p^\star),
\label{eq:judge2}
\end{equation}
and as $\beta\to0$ it converges to $p^\star-q=(\varepsilon_D-\varepsilon_M)(1-2q)/(1-2\varepsilon_M)$.
\end{corollary}

Both optima leave the reference in proportion to $1-2\varepsilon_M$, single comparison toward the mode and sibling pairing toward $q$.
When the reference is narrower than the data, a more accurate judge therefore narrows single comparison and widens sibling pairing, a sign reversal that judge exploitation would not produce.
The weight $w_M$ multiplies the sibling drive in \eqnref{eq:judge2}, so at the same $\beta$ \method moves further from a narrow reference than the quadratic form.

\begin{proof}[Proof]

With independent flips, the judge reports outcome 1 for a generated video with probability $\tilde p=\varepsilon_M+(1-2\varepsilon_M)p$ and for the recorded video with probability $\tilde q=\varepsilon_D+(1-2\varepsilon_D)q$.
Under single comparison, the expected reward of a generation with true outcome 1 minus that with outcome 0 equals $(1-2\varepsilon_M)(2\tilde q-1)=(1-2\varepsilon_M)(1-2\varepsilon_D)(2q-1)$.
A linear objective with KL penalty is maximized by tilting $P_{\mathrm{ref}}$ by the exponential of the expected reward divided by $\beta$, which gives \eqnref{eq:judge1}.
Under sibling pairing with $\lambda=1$, the siblings and the record are read through the same flips, so \propref{prop:pairing} applies to the reported outcomes and the objective is $-D_M(\tilde q\Vert\tilde p)+\mathrm{const}$ with $\partial\tilde p/\partial p=1-2\varepsilon_M$.
Over the two cells, its derivative with respect to $p$ is $(1-2\varepsilon_M)[w_M(\tilde p)+w_M(1-\tilde p)](\tilde q-\tilde p)$, and the derivative of $\mathrm{KL}(P_h\Vert P_{\mathrm{ref}})$ is $\logit p-\logit p_{\mathrm{ref}}$, so equating the two gives \eqnref{eq:judge2}.
As $\beta\to0$ the bracket stays bounded and positive, which forces $\tilde p^\star=\tilde q$ and $p^\star-q=(\varepsilon_D-\varepsilon_M)(1-2q)/(1-2\varepsilon_M)$.
For the energy form with continuous readouts, the same conclusion holds when both kinds of videos receive independent Gaussian readout noise from one distribution, because the Gaussian characteristic function has no zeros, so convolution is injective and the energy distance after noise vanishes only when it vanishes before noise.
\end{proof}

\subsection{Scope of the Results}

The propositions concern one history.
With parameters shared across histories, the expected update is $\E_h\nabla F(P_h)$, which for \method equals $-\E_h\nabla D_M(Q_h\Vert P_h)$, so the global maximizer is the data distribution whenever the model family can represent it, and otherwise the model closest to it in average divergence.
The distributions are those of the training sampler, so a deterministic sampler at deployment is guaranteed the same marginals only for the base model.
For \method, group size preserves the unregularized expected objective. The KL penalty changes the objective optimized by the finite-budget PPO updates.
\paragraph{Fixed kernels.}
For a kernel $\kappa$ that does not depend on the group, the proof of \lemref{lem:unbiased} applies with $\pi_P(a)=\E\,\kappa(a,\varphi(X))$, and (A1) gives $\E[\kappa_{gy}\omega_g\mid a_g=a]=\pi_Q(a)\,w_M(\pi_P(a))$ with $\pi_Q(a)=\E\,\kappa(a,\varphi(Y))$.
The expected reward of a generation at readout $a$ is therefore $w_M(\pi_P(a))(\pi_Q(a)-\lambda\pi_P(a))$, which vanishes for all $a$ when $P_h=Q_h$ and $\lambda=1$, so the data distribution is a stationary point.
For $M=0$ the reward is $\kappa(a_g,a_y)-\frac{\lambda}{G-1}\sum_{j\neq g}\kappa(a_g,a_j)$, and the symmetry argument of \appref{app:energy} gives the objective $\E\,\kappa(X,Y)-\frac{\lambda}{2}\E\,\kappa(X,X')$, which for $\lambda=1$ equals $-\frac12\mathrm{MMD}^2_\kappa(P_h,Q_h)+\mathrm{const}$ and has the unique maximizer $P_h=Q_h$ for a characteristic kernel such as the Gaussian kernel on $\R^m$~\cite{gretton2012kernel}.
A kernel whose bandwidth is computed from the current group, such as the median sibling distance, makes $\kappa_{gj}$ depend on all siblings and breaks the unbiasedness of $\omega_g$ and $\rho_g$, which is why \method fixes the bandwidth before training.

\subsection{The Quadratic Member in Energy Form}
\label{app:energy}

With the kernel induced by the Euclidean readout distance $d$, the quadratic member $M=0$ takes the energy form, which for a pair is $r(x)=-d(x,y)+\lambda\,d(x,x')$ and for a group is
\begin{equation}
r^0_g=-d(a_g,a_y)+\frac{\lambda}{G-1}\sum_{j\neq g}d(a_g,a_j),
\label{eq:m1e}
\end{equation}
with the leave-one-out baseline of \eqnref{eq:m2} computed from \eqnref{eq:m1e}.
For distributions $P,Q$ on $\R^m$ with finite first moments and independent $X,X'\sim P$ and $Y,Y'\sim Q$, define the self-dispersion and the energy distance
\begin{equation}
  S(P)=\E\,d(X,X'),\quad \ED(P,Q)=2\E\,d(X,Y)-S(P)-S(Q),
  \label{eq:ed}
\end{equation}
which is nonnegative and zero only for $P=Q$~\cite{szekely2013energy}.
The statements below are the energy-form counterparts of \propref{prop:pairing} and \propref{prop:group}, with objectives $F_0=-\tfrac12\ED(P_h,Q_h)-\tfrac12(1-\lambda)S(P_h)$ for an independent partner and $F_0-\tfrac{\lambda}{2}\Gamma_u$, $\Gamma_u=\tfrac12\E_{u,u'}\ED(P_{h,u},P_{h,u'})$, for a partner with shared $u$.

\paragraph{Independent partner.}
The reward is $-d(X,Y)+\lambda\,d(X,X')$, with $Y\sim Q_h$ and $X'\sim P_h$ treated as constants.
By (A1), $Y$ is independent of $\tau$, and applying \eqnref{eq:score} conditionally on $Y$ gives the term $\nabla(-\E\,d(X,Y))$.
For the sibling term, write $\Phi(\theta_1,\theta_2)=\E_{X\sim P_{\theta_1},X'\sim P_{\theta_2}}d(X,X')$, which is symmetric in its arguments and satisfies $S(P_h)=\Phi(\theta,\theta)$.
\eqnref{eq:score} differentiates only the first argument, and by symmetry $\nabla_{\theta_1}\Phi=\frac12\nabla S(P_h)$ at $\theta_1=\theta_2=\theta$.
Hence $F=-\E\,d(X,Y)+\frac{\lambda}{2}S(P_h)$.
Substituting $\E\,d(X,Y)=\frac12[\ED(P_h,Q_h)+S(P_h)+S(Q_h)]$ from \eqnref{eq:ed} yields $F_0$ with constant $-\frac12 S(Q_h)$.
For $\lambda=1$, $F=-\frac12\ED(P_h,Q_h)+\mathrm{const}$, and since the energy distance between distributions on $\R^m$ with finite first moments is nonnegative and vanishes only for identical distributions~\cite{szekely2013energy}, the unique maximizer is $P_h=Q_h$.
For $\lambda\neq1$, the extra term $-\frac12(1-\lambda)S(P_h)$ rewards shrinking ($\lambda<1$) or expanding ($\lambda>1$) the model's dispersion, so $P_h=Q_h$ is not stationary in general.

\paragraph{Partner with shared randomness.}
Let $u$ be the shared random numbers and write $X=g_\theta(h,u,\xi_1)$ and $X'=g_\theta(h,u,\xi_2)$ with independent $\xi_1,\xi_2$.
Because $u$ does not depend on $\theta$, $\log\pi_\theta(\tau)$ contains only the transitions after $u$, and applying the previous argument conditionally on $u$ gives $F=-\E\,d(X,Y)+\frac{\lambda}{2}\E_u S(P_{h,u})$.
For independent $u,u'$,
\begin{equation}
\begin{aligned}
&\E_{u,u'}\big[2\,\E\,d(X_u,X_{u'})-S(P_{h,u})-S(P_{h,u'})\big]\\
&\qquad=2S(P_h)-2\,\E_uS(P_{h,u}),
\end{aligned}
\end{equation}
because two generations with independent $u$ and $u'$ are independent draws from $P_h$.
The left side equals $2\Gamma_u$, so $\E_uS(P_{h,u})=S(P_h)-\Gamma_u$, and substitution gives $F_0-\tfrac{\lambda}{2}\Gamma_u$.
Nonnegativity of $\Gamma_u$ follows from that of the energy distance.

\paragraph{Partner from other histories.}
If $X'$ is drawn from the pooled generations $\bar P$ of other histories and $\bar P$ is held fixed, $\E\,d(X,X')=\int P_h(dx)\int\bar P(dx')\,d(x,x')$ is linear in $P_h$, and so is the whole objective.
Since $\bar P$ moves with the parameters, the result characterizes fixed points of the update.

\paragraph{Binary outcomes.}
With $d(a,b)=\ind[a\neq b]$, which is the one-hot Euclidean distance up to the factor $\sqrt2$, we have $S(P_h)=2p(1-p)$ and $\E\,d(X,Y)=p(1-q)+q(1-p)$.
Then $F=-p-q+2pq+\lambda p(1-p)$, whose derivative $(2q-1)+\lambda(1-2p)$ vanishes at \eqnref{eq:pstar}.
For $\lambda>0$, $F$ is strictly concave in $p$, so this point is the unique maximizer after clipping to $[0,1]$, and for $\lambda=0$ the linear objective is maximized at the mode.
For $\lambda=1$, $F=-(p-q)^2-q(1-q)$ is the negative Brier risk, the $M=0$ case of \propref{prop:pairing}.

For shared noise, let $k(u)=P_{h,u}(1)$ and $V=\mathrm{Var}_u\,k(u)$.
Free replays disagree with probability $D_{\mathrm{free}}=2p(1-p)$ and replays with tied $u$ with probability $D_{\mathrm{tied}}=2\E_u[k(1-k)]=2p(1-p)-2V$, so $\Gamma_u=D_{\mathrm{free}}-D_{\mathrm{tied}}=2V$ and, for $\lambda=1$, $F=-(p-q)^2-V+\mathrm{const}$.
If training moves the model along a parameter direction $c$, the stationary point satisfies
\begin{equation}
p^\star-q=-\tfrac12\,\frac{\partial V/\partial c}{\partial p/\partial c}.
\label{eq:sharedbias}
\end{equation}
For $k(u)=\sigma(c+a u)$ with $u\sim\mathcal N(0,1)$, trainable $c$ and $q=0.65$, the stationary points for $a=1,2,4$ are $0.699$, $0.778$ and $0.905$ ($0.677$, $0.738$ and $0.858$ for the reward of \method with $M=G-2$, $G=8$), so the bias points toward the mode and grows with the influence $a$ of the shared noise, which is what the experiments measure through $\Gamma_u$.
A model that could make the outcome independent of $u$ without changing $p$ would remove this bias, and the KL penalty, the low-rank parameterization and the finite number of updates prevent this in practice.

For partners from other histories with $\lambda=1$, the gradient for history $h$ is $2(q_h-\bar p)$ and does not depend on $p_h$.
At a fixed point with pooled frequency $t=\bar p$, histories with $q_h>t$ move to $p_h=1$ and those with $q_h<t$ to $p_h=0$, while $t$ is determined self-consistently by these limits.
Only when all histories share the same $q_h$ does this coincide with sibling pairing.

\paragraph{Group advantages for the energy form.}
The baseline $b_g$ depends on $(h,y,x_{-g})$, which is independent of $\tau_g$ given $h$, so $\E[b_g\nabla\log\pi_\theta(\tau_g)]=\E[b_g]\,\E[\nabla\log\pi_\theta(\tau_g)]=0$.
The scale $c$ does not depend on the group.
Each of the $G-1$ sibling terms $\frac{1}{G-1}d(x_g,x_j)$ contributes $\frac{1}{G-1}\cdot\frac12\nabla S(P_h)$ by the argument above, so the sibling terms sum to $\frac12\nabla S(P_h)$, which is case~1 with $\lambda=1$.
The leave-one-out baseline is the mean of \eqnref{eq:m1e} over the group without $x_g$, so it depends only on $x_{-g}$ and $y$.
For the mean of the other rewards, let $B_g=\frac1{G-1}\sum_{j\neq g}r_j$.
Every $r_j$ contains the term $\frac{1}{G-1}d(x_j,x_g)$, which depends on $x_g$, while its remaining terms do not and contribute zero.
Therefore $\E[B_g\nabla\log\pi_\theta(\tau_g)]=\frac{1}{(G-1)^2}\sum_{j\neq g}\frac12\nabla S(P_h)=\frac{1}{G-1}\cdot\frac12\nabla S(P_h)$.
Subtracting this from the sibling contribution $\frac12\nabla S(P_h)$ of $r_g$ leaves $\big(1-\frac{1}{G-1}\big)\frac12\nabla S(P_h)$, so $\lambda_{\mathrm{eff}}=\frac{G-2}{G-1}$.
The GRPO form satisfies $r_g-\bar r=\frac{G-1}{G}(r_g-B_g)$ and differs only by a positive factor. Mean baselines are unbiased for rewards that do not depend on other samples~\cite{ahmadian2024rloo,liu2025drgrpo}, and the bias here arises only because sibling-paired rewards do.
For group standard-deviation normalization and a binary readout with $n_1$ generations of outcome 1 and $n_0$ of outcome 0, all generations with the same outcome receive the same reward, $r(1)=-\ind[y\neq1]+\lambda n_0/(G-1)$ and $r(0)=-\ind[y\neq0]+\lambda n_1/(G-1)$.
If only one outcome occurs, the standardized advantages are zero for every $\lambda$.
Otherwise the standardized vector depends only on $(n_0,n_1)$ and the sign of $r(1)-r(0)$.
For $y=1$, $r(1)-r(0)=1+\lambda(n_0-n_1)/(G-1)\geq1-\lambda(G-2)/(G-1)$, which is positive for $\lambda<(G-1)/(G-2)$ and has the same sign as for $\lambda=0$, and $y=0$ is symmetric.
The advantages therefore coincide with those of $\lambda=0$ in every group.
For $\lambda=0$ and $y=1$, the outcome-1 generations receive $\sqrt{n_0/n_1}$ and the others $-\sqrt{n_1/n_0}$, and the logit score of each generation is $R_g-p$, so the update along the logit is $\sqrt{n_0n_1}/G$, with the opposite sign for $y=0$.
Its expectation $(2q-1)\E\sqrt{n_0n_1}/G$ is strictly positive for $q>\tfrac12$ and $0<p<1$.
Without KL regularization, the attracting limit is $p\to1$ for every $G\geq2$, and $q<\tfrac12$ is symmetric.

\section{Training and Evaluation Details}
\label{app:setup}

\begin{algorithm}[t]
\caption{One update of \method}
\label{alg:sibpo}
\small
\begin{algorithmic}[1]
\Require histories $\{h_p,y_p\}_{p=1}^{P}$, group size $G\geq3$, readout $\varphi$, kernel $\kappa$, order $M=G-2$, running scale $c$
\For{$p=1,\dots,P$}
  \State encode $h_p$ once and share the encoding across the group
  \For{$g=1,\dots,G$}
    \State sample $x_{p,g}$ with independent initial and step noise and store $\log\pi_\theta(\tau_{p,g})$
  \EndFor
  \State $a_{p,g}\gets\varphi(x_{p,g})$, $a_{p,y}\gets\varphi(y_p)$, $K_{jk}\gets\kappa(a_{p,j},a_{p,k})$
  \State $\omega_{p,g},\rho_{p,g}\gets$ symmetric polynomials of $\{1-K_{gj}\}_{j\neq g}$
  \State $r_{p,g}\gets \kappa(a_{p,g},a_{p,y})\,\omega_{p,g}-\rho_{p,g}$ \Comment{\eqnref{eq:m1}}
  \State $b_{p,g}\gets$ mean of \eqnref{eq:m1} over the group without $g$ \Comment{\eqnref{eq:m2}}
  \State $A_{p,g}\gets (r_{p,g}-b_{p,g})/c$
\EndFor
\State update LoRA weights with the PPO clipped objective on $A_{p,g}$ and a KL penalty
\State update the running scale $c$ with the values $|r-b|$ of this batch
\end{algorithmic}
\end{algorithm}

\paragraph{Procedure.}
\algoref{alg:sibpo} summarizes one update.
The symmetric polynomials $e_0,\dots,e_M$ of the $G-1$ miss probabilities follow from multiplying the linear factors $(1+(1-K_{gj})t)$, which costs $O(G^2)$ per generation and $O(G^3)$ per group, under $0.05$ GPU hours per run, and leaving out one sibling divides out one factor, so $\rho$ and the baseline cost the same order.
For $M=0$ the reward reduces to the additive energy form \eqnref{eq:m1e}.
All baselines use the same group sampling, PPO clipping and KL penalty, and differ only in the reward and advantage computation, with the exceptions noted below.

\paragraph{Configuration.}
All post-training methods share one configuration within a model family, and the four video models use the same values.
Training uses LoRA rank 64, $G=8$ generations for each of $P=4$ histories per update, 150 updates, a resolution of $256\times448$ with 25 frames, a 10-step stochastic sampler and a PPO clip range of 0.2.
Each method selects its KL weight from $\{0.01,0.04\}$ and its learning rate from $\{1,3\}\times10^{-5}$ once on the validation split, every configuration is trained with three seeds, and each run draws $150\times32=4{,}800$ generations.
A \method run takes 21.6, 18.7, 28.4 and 126.0 GPU hours on Cosmos-Predict2.5-2B, Wan2.2-TI2V-5B, CogVideoX1.5-5B-I2V and Cosmos-Predict2.5-14B.
Training runs on four A100-80GB GPUs in bf16.
Evaluation follows the official resolution and duration of each benchmark, which is about 480p for the video benchmarks, with the training sampler and step count.
Training data cover five mechanism families: the public PISA simulated drop videos, the Physion training split, and throws, bounces and collisions of rigid objects generated with the public Kubric simulator~\cite{greff2022kubric}, 4{,}000 histories per Kubric family with the restitution and initial velocity drawn per scene. Splits are by source scene, and $10\%$ of training scenes are held out for selecting the learning rate and KL weight.
With $P=4$ histories per update, each run draws 600 histories from the training pool and uses each at most once, so a Physion all-but-one run removes on the order of 15 histories of the held-out scenario. The held-out runs repeat the full training of every method on the mixture without the falling family, and on the mixture without each of the eight Physion scenarios in turn, with the configuration selected once on the full mixture.
No test image of any benchmark is used for training or selection.
The experiments reported here took about $81{,}000$ A100 GPU hours, of which $15{,}500$ were training, mostly the 240 Physion all-but-one runs and the 30 Cosmos-14B runs, and the rest evaluation, dominated by the eight Physion continuations per test video, the 128 generations per PisaBench frame, the 64 generations per Kubric test scene and the 14B evaluations.
The EBFT, success-count, $M=0$ and sibling-weight runs of \tabref{tab:family} add $336.9$ training GPU hours on Cosmos-Predict2.5-2B, from three seeds each of success counts, $M=0$ and $\lambda=0.5$ and $1.5$ ($259.2$) and of EBFT ($77.7$, at $25.9$ per run because it encodes every generation with the frozen generator), and their evaluation follows the same protocol.
The rows of EBFT and of \method with $M=0$ in \tabref{tab:main}, \tabref{tab:scope}, \tabref{tab:families-full} and \tabref{tab:14b-full} reuse these Cosmos-Predict2.5-2B checkpoints and add $1{,}725.2$ training GPU hours: $583.2$ for the 27 held-out runs of $M=0$ on Cosmos-Predict2.5-2B, and three seeds each of $M=0$ and EBFT on Wan2.2 ($56.1$ and $67.3$), CogVideoX ($85.2$ and $102.2$) and Cosmos-Predict2.5-14B ($378.0$ and $453.3$). An $M=0$ run costs as much as a \method run, and an EBFT run costs $25.9/21.6$ times as much in every family.

\begin{table}[t]
  \centering
  \caption{\textbf{Reward-family members across group sizes and from a collapsed start on Cosmos-Predict2.5-2B.}
  Left: PisaBench KS distance at 32 generations per update (default $G=8$). Right: IQR ratio when training continues from the Flow-GRPO checkpoint (IQR ratio $0.55$).}
  \label{tab:group}
  \tablestyle
  \begin{tabularx}{\linewidth}{l||YYY|YYY}
    \thickhline
    \multirow{2}{*}{\hd{Reward}} & \multicolumn{3}{c|}{\hd{KS\dn\ at group size}} & \multicolumn{3}{c}{\hd{IQR\one\ after restart}} \\
    \cline{2-7}
     & $G=4$ & $G=8$ & $G=16$ & 10 upd. & 25 upd. & 50 upd. \\
    \hline\hline
Flow-GRPO (single comparison) & 0.326 & 0.329 & 0.334 & 0.55 & 0.54 & 0.53 \\
Success count~\cite{tajwar2026maxrl} & 0.287 & 0.249 & 0.244 & 0.61 & 0.67 & 0.74 \\
\method, $M=0$ & 0.259 & 0.256 & 0.254 & 0.59 & 0.62 & 0.67 \\
\hline
\method & \textbf{0.246} & \textbf{0.241} & \textbf{0.239} & \textbf{0.63} & \textbf{0.71} & \textbf{0.79} \\
    \thickhline
  \end{tabularx}
\end{table}

\paragraph{Group size and restart.}
The group-size runs of \tabref{tab:group} (left) keep 32 generations per update by setting $P=32/G$, so $G=4$, $8$ and $16$ use $P=8$, $4$ and $2$ histories per update with the same 150 updates, learning rate and KL weight as the default, and the mean-of-others and group standard-deviation rows of \tabref{tab:pairing} follow the same rule.
Success counts use $M=G-1$ and \method uses $M=G-2$.
The restart runs of \tabref{tab:group} (right) load each of the three Flow-GRPO checkpoints after 150 updates, continue for 50 updates with the reward of the respective member and the same optimizer state reset, and evaluate PisaBench after 10, 25 and 50 updates.
These runs add $604.1$ training GPU hours, $517.8$ for the 24 group-size runs at $G=4$ and $16$ and $86.3$ for the 12 restart runs.

\paragraph{Readouts and judges.}
The training readout of falling and thrown objects tracks the object with CoTracker3~\cite{karaev2025cotracker3} and intersects the track with a ground line estimated in the first frame.
For falling objects the outcome cell is the landing frame.
For Kubric families it is the landing position along the throw direction, the number of bounces before rest or the exit direction after the collision, where position and direction fall into eight cells bounded by the octiles of the recorded training videos of the family, computed once before training.
The training readout of Physion contact tests the adjacency of SAM~2~\cite{ravi2025sam2} masks of the two queried objects in the last frame, with a separate class for videos in which either object is lost.
Evaluation always uses the official judges, namely the PISA mask pipeline, the Physion readout protocol of SlotFormer~\cite{wu2023slotformer}, PAWEval and the vision-language readout of CaliBench.
Readout errors are measured on 200 held-out recorded videos against simulator ground truth and on 200 generations per checkpoint against the agreement of two annotators.

\paragraph{Compared methods.}
RLVR-World uses the negative L1 plus LPIPS distance to the recorded frames with GRPO advantages, which on flow models amounts to Flow-GRPO with this reward.
Flow-GRPO uses the negative training-readout distance with group mean and standard-deviation normalization.
The three PISA-ORO variants backpropagate the segmentation, flow and depth rewards through the decoder as in the original work, which explains their higher memory.
DynamicsBoost builds preference pairs from continuations conditioned on recorded prefixes of different lengths and trains with its asymmetric DPO loss.
DiverseGRPO adds its cluster-based diversity reward, computed by spectral clustering of frozen video-encoder features of the group, with the default weight of the original work.
DPH-RL replaces the reverse KL penalty of Flow-GRPO by the forward KL to samples of the base model.
EBFT~\cite{jelassi2026ebft} matches mean features of generated and recorded futures with its alignment-minus-diversity reward on features of a frozen copy of the generator, and success-count normalization~\cite{tajwar2026maxrl} divides the record-agreement reward by the number of agreeing generations, both with the group sampling and readouts of \method.
\method with $M=0$ uses \eqnref{eq:m1} with $\omega_g=1$, which is the quadratic score of the readout cells, together with the leave-one-out baseline and running scale of \eqnref{eq:m2}, so it shares every setting of \method except the weight.
For the decoded robot futures, $\sigma$ is the median L2 distance between pairs of base-model generations of the same training history, computed once on 200 training histories before post-training.
SFT continues flow-matching training~\cite{lipman2023flow,liu2023rectified} on recorded futures for the same number of GPU hours.
On the token model iVideoGPT, SFT continues likelihood training, DynamicsBoost applies its asymmetric DPO loss to token log-probabilities, DiverseGRPO and DPH-RL modify the GRPO update of RLVR-World as above, and Flow-GRPO and PISA-ORO do not apply, since they require a stochastic flow sampler or a differentiable decoder path.

\paragraph{Statistics.}
Each benchmark is first aggregated within its natural unit, which is the first frame for PisaBench, the scene for Kubric, PAWBench and CaliBench and the test video for Physion, with the eight Physion scenarios weighted equally.
Intervals for differences between methods resample units and training seeds in two levels with 10{,}000 bootstrap draws on unrounded values, and $\Gamma_u$ intervals resample the 50 PisaBench frames.

\section{Benchmarks and Protocols}
\label{app:bench}

\begin{figure*}[t]
  \centering
  \includegraphics[width=\textwidth]{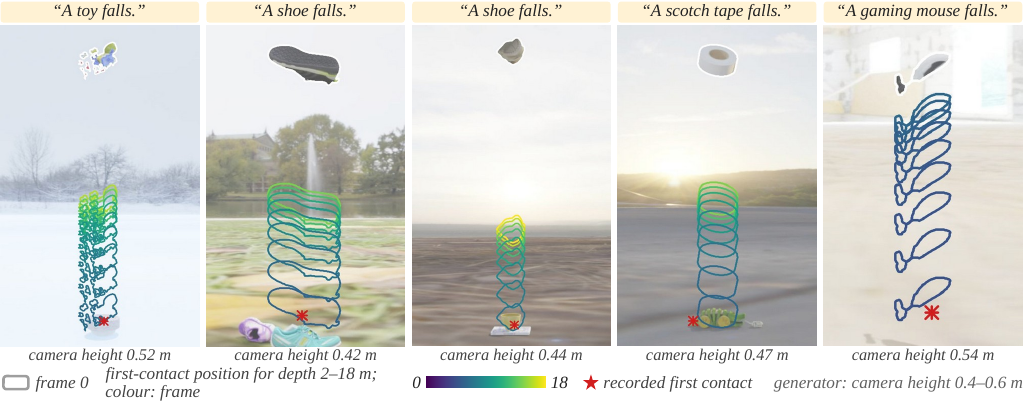}
  \caption{\textbf{Why a PisaBench first frame admits many landing times.}
  For each first frame, the curves mark where the object would first touch the ground if it were placed at depths between 2 and 18\,m with the camera height of the generator, and the color gives the frame of contact.
  The recorded video (star) realizes one of these depths, so the analytic distribution of landing times has a width that the frame cannot resolve.}
  \label{fig:pisa-ambiguity}
\end{figure*}

\paragraph{PisaBench distribution protocol.}
The protocol of PISA~\cite{li2025pisa} uses simulated drops in which the image of an object at a given height is consistent with a range of depths, because the depth of the object cannot be read from the frame and the camera height varies between 0.4 and 0.6\,m.
\figref{fig:pisa-ambiguity} shows how a single first frame maps to a family of possible first-contact positions and frames.
For each of the 50 first frames, the simulator provides 1,000 Monte Carlo landing times quantized to frames, and each model generates 128 futures.
The KS statistic compares the two samples, and the reported KS distance is its mean over frames.
Because a deviation in location alone also raises the KS statistic, we report the IQR ratio next to it, and only the IQR ratio indicates concentration.
The readout dispersion ratio $S$ divides the mean pairwise distance of generated landing readouts by that of simulated ones and corresponds directly to $S(P_h)/S(Q_h)$.
Most models are rejected by the KS test on most frames, since 128 generations resolve very small differences, so the rejection count carries little information and we report the mean statistic.

\paragraph{Physion.}
The 1,200 test videos cover eight scenarios with balanced contact outcomes.
From the visible prefix, each model generates eight continuations, and the official readout returns a contact probability per continuation.
The video-level prediction is the mean probability, accuracy thresholds it at $0.5$, the Pearson correlation with the mean human response is computed within each scenario and averaged, and Cohen's $\kappa$ is averaged over individual participants.
The human response measures how ambiguous a prefix appears to observers and differs from the true probability, so we also report the fair Brier score of the eight binary readouts against the true outcome, the only proper score among these metrics.
Pearson correlation and $\kappa$ can increase when predictions become sharper while keeping their ranking, so they are read together with the Brier score.

\paragraph{Transfer benchmarks.}
PAWBench~\cite{pu2026pawbench} generates 50 videos per scene for 25 calibration scenes and reports the total variation distance to the reference outcome distribution, averaged over scenes whose outcomes pass the readability threshold, together with the coverage of legal outcomes on 25 further scenes.
CaliBench~\cite{sadeghi2026calibench} generates 32 videos for each of nine scenes with analytic outcome distributions and reports mnTV, which subtracts the finite-sample expectation of the TVD under a calibrated model and penalizes unreadable generations.
Neither benchmark shares a mechanism with the training data.

\paragraph{Kubric distributional split.}
The 150 test scenes (50 each for throws, bounces and collisions) come from source scenes that never occur in training.
For each scene we keep the rendered first frame and re-run the simulator 500 times with the restitution, mass and initial speed drawn from the ranges of the training generator, conditioned on the visible initial pose, so the reference distribution covers exactly what the frame leaves open.
The outcome is the landing position along the throw direction, the number of bounces before rest and the exit direction after the collision.
The evaluation readout segments the object with SAM~2 and intersects its mask with the ground plane, which differs from the CoTracker3 readout used for training, and each family enters the mean KS distance with equal weight.
This split is ours and is reported in its own column, separately from the public benchmarks.

\paragraph{Spread-matched deployment.}
For the base model and Flow-GRPO we search the guidance scale over $\{0.5,0.75,1\}\,w_0$ and the sampler noise multiplier over $\{1,1.2,\dots,2\}$ on the validation scenes and keep the setting whose pooled readout dispersion ratio $S$ is closest to one, which is $0.75\,w_0$ with multiplier $1.6$ for Flow-GRPO and $0.75\,w_0$ with multiplier $1.2$ for the base model.
The same setting is then used on every benchmark.

\paragraph{Single-video quality.}
PisaBench trajectory metrics compare object masks and trajectories of one generation with the recorded video on 361 real videos.
Physics-IQ~\cite{motamed2025physicsiq} compares generated and real motion in 198 real image-to-video scenes with four runs each, and VBench-I2V~\cite{huang2025vbenchpp} measures consistency, motion smoothness, dynamic degree and image quality on its official suite.
A model that suppresses motion can score high on consistency, so the dynamic degree is always reported with it.

\paragraph{RT-1 and Real2Sim.}
For iVideoGPT~\cite{wu2024ivideogpt} we follow RLVR-World~\cite{wu2025rlvrworld}: multi-step prediction on RT-1~\cite{brohan2023rt1} is scored by MSE, PSNR, SSIM and LPIPS, and Real2Sim evaluation rolls out policies in the world model, records task completion with a criterion fixed before evaluation, and compares the resulting success rates with real-robot success rates.
The original protocol covers four RT-1 and RT-1-X checkpoints on six drawer tasks. We extend it to the five open policies with real-robot evaluations in SIMPLER~\cite{li2024simpler}, namely RT-1 at the beginning, at $15\%$ and at the end of training, RT-1-X and Octo-Base~\cite{octo2024}, and to the ten Google Robot task variants with reported real-robot success rates, namely picking a can in three orientations, moving an object near another and opening and closing the top, middle and bottom drawers. This gives 50 policy-task pairs, each with 20 world-model rollouts from fixed initial frames, and the real-robot rates are taken from SIMPLER without new robot trials.
Each history receives 100 generations, of which the first 32 measure dispersion as the mean pairwise L1 distance of decoded futures.

\section{Other Model Families, Scale and Policy Evaluation}
\label{app:ar}

\begin{table}[t]
  \centering
  \caption{\textbf{Trained-mechanism comparison on two further model families and at 14B scale.}
  Same methods, data, 150 updates, three seeds and protocol as \tabref{tab:main}, plus EBFT and the quadratic member of the family under the same budget. KS and IQR ratio (ideally 1) are PisaBench, Brier is Physion. Held-out and single-video metrics are in \tabref{tab:families-full} and \tabref{tab:14b-full}.}
  \label{tab:scope}
  \tablestyle
  \begin{tabularx}{\linewidth}{l||YYY|YYY|YYY}
    \thickhline
    \multirow{2}{*}{\hd{Method}} & \multicolumn{3}{c|}{\hd{Wan2.2-5B}} & \multicolumn{3}{c|}{\hd{CogVideoX-5B}} & \multicolumn{3}{c}{\hd{Cosmos-14B}} \\
    \cline{2-10}
     & KS\dn & IQR & Brier\dn & KS\dn & IQR & Brier\dn & KS\dn & IQR & Brier\dn \\
    \hline\hline
Base model & 0.653 & 0.64 & 0.269 & 0.537 & 0.75 & 0.254 & 0.537 & 0.59 & 0.262 \\
\cdashline{1-10}
SFT & 0.348 & 0.74 & 0.264 & 0.294 & 0.85 & 0.251 & 0.296 & 0.72 & 0.255 \\
\hline\grp{10}{Rewards against the recorded future}\\
RLVR-World & 0.371 & 0.51 & 0.287 & 0.312 & 0.61 & 0.279 & 0.352 & 0.47 & 0.281 \\
Flow-GRPO & 0.368 & 0.50 & 0.290 & 0.309 & 0.62 & 0.276 & 0.361 & 0.45 & 0.286 \\
ORO (Seg) & 0.355 & 0.55 & 0.284 & 0.300 & 0.65 & 0.272 & 0.338 & 0.51 & 0.278 \\
ORO (Flow) & 0.364 & 0.52 & 0.288 & 0.306 & 0.63 & 0.276 & 0.349 & 0.48 & 0.283 \\
ORO (Depth) & 0.360 & 0.53 & 0.283 & 0.303 & 0.64 & 0.270 & 0.344 & 0.49 & 0.279 \\
DynamicsBoost & 0.329 & 0.62 & 0.270 & 0.279 & 0.74 & 0.258 & 0.301 & 0.60 & 0.262 \\
\hline\grp{10}{Diversity-preserving post-training}\\
DiverseGRPO & 0.362 & 0.53 & 0.281 & 0.304 & 0.64 & 0.268 & 0.348 & 0.48 & 0.274 \\
DPH-RL & 0.318 & 0.66 & 0.262 & 0.271 & 0.79 & \textbf{0.250} & 0.289 & 0.63 & 0.256 \\
\hline\grp{10}{Sibling pairing with a quadratic score}\\
EBFT~\cite{jelassi2026ebft} & 0.309 & 0.69 & 0.263 & 0.277 & 0.78 & 0.253 & 0.283 & 0.66 & 0.257 \\
\method, $M=0$ & 0.301 & 0.73 & 0.256 & 0.270 & 0.83 & 0.252 & 0.276 & 0.69 & 0.248 \\
\hline\hline
\rowcolor{rowblue}\method (ours) & \textbf{0.281} & \textbf{0.82} & \textbf{0.251} & \textbf{0.262} & \textbf{0.87} & 0.251 & \textbf{0.254} & \textbf{0.80} & \textbf{0.243} \\
    \thickhline
  \end{tabularx}
\end{table}

\begin{table}[t]
  \centering
  \caption{\textbf{Robot policy evaluation with an autoregressive world model.}
  iVideoGPT~\cite{wu2024ivideogpt} on RT-1. Disp.\ is the mean pairwise L1 distance of 32 generations, and Real2Sim compares world-model with real-robot success rates~\cite{li2024simpler}, with MAE $\times10^{-2}$. The variant row uses the L1+LPIPS distance.}
  \label{tab:ar}
  \tablestyle
  \begin{tabularx}{\linewidth}{l||YYY|YYY}
    \thickhline
    \multirow{2}{*}{\hd{Method}} & \multicolumn{3}{c|}{\hd{RT-1 ($\times10^{-2}$)}} & \multicolumn{3}{c}{\hd{Real2Sim}} \\
    \cline{2-7}
     & MSE\dn & LPIPS\dn & Disp. & MAE\dn & $\rho$\up & Brier\dn \\
    \hline\hline
Base model & 0.659 & 14.8 & 3.53 & 17.2 & 0.71 & 0.214 \\
\cdashline{1-7}
SFT & 0.602 & 14.2 & 3.29 & 15.4 & 0.74 & 0.207 \\
RLVR-World (released)~\cite{wu2025rlvrworld} & \textbf{0.486} & \textbf{13.4} & 2.41 & 14.6 & 0.76 & 0.219 \\
RLVR-World (retrained) & 0.491 & 13.6 & 2.47 & 14.3 & 0.76 & 0.217 \\
DynamicsBoost~\cite{li2026dynamicsboost} & 0.561 & 14.0 & 3.02 & 13.8 & 0.80 & 0.209 \\
DiverseGRPO~\cite{liu2026diversegrpo} & 0.509 & 13.7 & 2.66 & 14.1 & 0.76 & 0.213 \\
DPH-RL~\cite{li2026dphrl} & 0.527 & 13.9 & 3.11 & 12.9 & 0.79 & 0.203 \\
\hline\hline
\rowcolor{rowblue}\method, L1+LPIPS distance & 0.532 & 13.8 & 3.18 & 12.2 & 0.80 & 0.199 \\
\rowcolor{rowblue}\method (ours) & 0.520 & 14.1 & 3.31 & \textbf{11.2} & \textbf{0.82} & \textbf{0.196} \\
    \thickhline
  \end{tabularx}
\end{table}

\begin{figure*}[t]
  \centering
  \includegraphics[width=\textwidth]{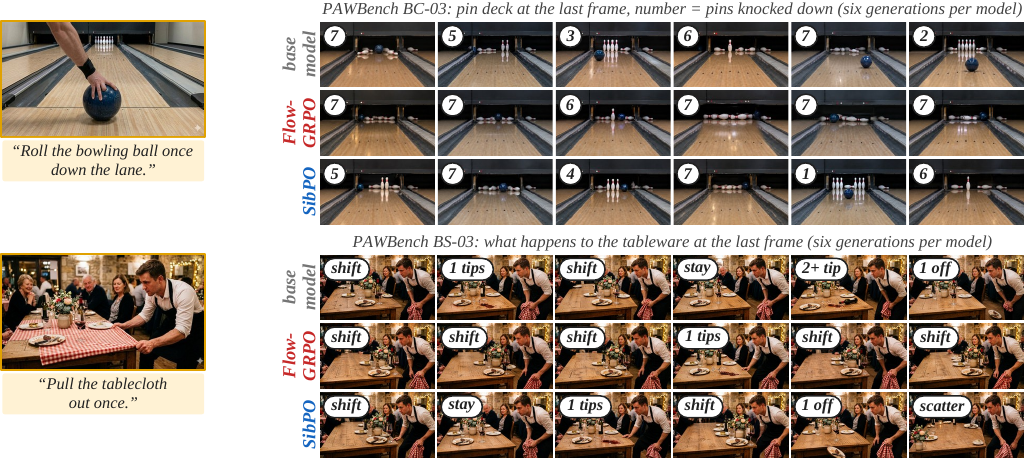}
  \caption{\textbf{Transfer to untrained mechanisms on PAWBench.}
  Six generations per model for a bowling scene, labeled with the number of knocked-down pins, and a tablecloth pull, labeled with the fate of the tableware.
  Flow-GRPO repeats a few outcomes more often than the base model, and \method keeps a spread similar to the base model.
  Input frames are the official PAWBench source images of scenes BC-03 and BS-03, and the corner mark is part of the distributed files.}
  \label{fig:paw}
\end{figure*}

\begin{table*}[t]
  \centering
  \caption{\textbf{Held-out and single-video metrics on Wan2.2-TI2V-5B and CogVideoX1.5-5B-I2V.}
  Complements \tabref{tab:scope} with the remaining metrics of \tabref{tab:main} under the same protocol. Pear.\ is the Physion correlation with humans, Acc.\ the Physion accuracy, L2 the PisaBench real-video trajectory error ($\times10^{-2}$), Dyn.\ the VBench-I2V dynamic degree, and PAWBench TVD ($\times100$) and CaliBench mnTV score held-out mechanisms. $^\ddagger$Our Kubric distributional split.}
  \label{tab:families-full}
  \tablestyle
  \begin{tabularx}{\textwidth}{l||YY|Y|Y|Y|Y|YY||YY|Y|Y|Y|Y|YY}
    \thickhline
    \multirow{3}{*}{\hd{Method}} & \multicolumn{8}{c||}{\hd{Wan2.2-TI2V-5B}} & \multicolumn{8}{c}{\hd{CogVideoX1.5-5B-I2V}} \\
    \cline{2-17}
     & \multicolumn{2}{c|}{Physion} & Kub.$^\ddagger$ & Pisa. & Ph.-IQ & VBench & PAW. & Cali. & \multicolumn{2}{c|}{Physion} & Kub.$^\ddagger$ & Pisa. & Ph.-IQ & VBench & PAW. & Cali. \\
    \cline{2-17}
     & Pear.\up & Acc.\up & KS\dn & L2\dn & Score\up & Dyn.\up & TVD\dn & mnTV\dn & Pear.\up & Acc.\up & KS\dn & L2\dn & Score\up & Dyn.\up & TVD\dn & mnTV\dn \\
    \hline\hline
Base model & 0.330 & 59.3 & 0.437 & 12.8 & 23.1 & 46.0 & 31.2 & 0.587 & 0.303 & \textbf{60.8} & 0.389 & 13.4 & 27.1 & 33.0& 29.9 & 0.641 \\
\cdashline{1-17}
SFT & 0.361 & 59.6 & 0.322 & 10.0 & 23.8 & 46.5 & \textbf{29.9} & 0.596 & 0.338 & 59.6 & 0.288 & 10.0 & 27.8 & 35.4& 31.1 & 0.642 \\
\hline\grp{17}{Rewards against the recorded future}\\
RLVR-World~\cite{wu2025rlvrworld} & 0.337 & 58.8 & 0.352 & 9.3 & 22.0 & 44.1 & 32.4 & 0.611 & 0.331 & 59.0 & 0.319 & 9.4 & 27.4 & 31.8& 32.5 & 0.618 \\
Flow-GRPO~\cite{liu2025flowgrpo} & 0.321 & 59.0 & 0.357 & 9.1 & 24.4 & 45.0 & 32.1 & 0.606 & 0.338 & 59.2 & 0.322 & 9.2 & 28.2 & 32.4& 32.0 & \textbf{0.611} \\
ORO (Seg)~\cite{li2025pisa} & 0.318 & 59.1 & 0.341 & 9.2 & 23.2 & 45.6 & 32.6 & 0.603 & 0.312 & 59.5 & 0.307 & 9.3 & 26.6 & 33.1& 32.7 & 0.622 \\
ORO (Flow)~\cite{li2025pisa} & 0.329 & 58.9 & 0.350 & 9.0 & 25.2 & 47.9 & 31.8 & 0.609 & 0.326 & 59.1 & 0.315 & 9.0 & \textbf{29.6} & 35.9& 31.6 & 0.626 \\
ORO (Depth)~\cite{li2025pisa} & 0.334 & 59.2 & 0.346 & \textbf{8.9} & 24.7 & 47.0 & 31.4 & 0.612 & 0.341 & 59.4 & 0.311 & \textbf{8.9} & 27.7 & 35.0& 31.2 & 0.620 \\
DynamicsBoost~\cite{li2026dynamicsboost} & 0.344 & 59.4 & 0.311 & 9.4 & \textbf{26.1} & \textbf{49.8} & 30.6 & 0.590 & 0.336 & 59.8 & 0.284 & 9.4 & 29.0 & \textbf{37.6}& 30.5 & 0.629 \\
\hline\grp{17}{Diversity-preserving post-training}\\
DiverseGRPO~\cite{liu2026diversegrpo} & 0.352 & 59.1 & 0.349 & 9.3 & 23.1 & 48.3 & 30.2 & 0.589 & 0.347 & 59.5 & 0.314 & 9.3 & 26.4 & 36.2& \textbf{29.8} & 0.613 \\
DPH-RL~\cite{li2026dphrl} & 0.348 & \textbf{59.7} & 0.304 & 9.4 & 21.2 & 46.4 & 30.4 & 0.593 & 0.344 & 60.3 & 0.279 & 9.5 & 25.6 & 34.0& 30.1 & 0.633 \\
\hline\grp{17}{Sibling pairing with a quadratic score}\\
EBFT~\cite{jelassi2026ebft} & 0.350 & 59.2 & 0.318 & 9.6 & 23.3 & 48.8 & 30.6 & 0.592 & 0.340 & 60.1 & 0.296 & 9.7 & 27.2 & 37.0 & 30.0 & 0.619 \\
\method, $M=0$ & 0.362 & 59.6 & 0.291 & 9.3 & 23.7 & 47.4 & 30.3 & 0.588 & 0.346 & 60.5 & 0.274 & 9.4 & 27.7 & 35.3 & 30.2 & 0.627 \\
\hline\hline
\rowcolor{rowblue}\method (ours) & \textbf{0.372} & 59.5 & \textbf{0.279} & 9.2 & 23.6 & 47.9 & 30.0 & \textbf{0.584} & \textbf{0.351} & 60.7 & \textbf{0.271} & 9.4 & 28.1 & 35.1& 30.3 & 0.624 \\
    \thickhline
  \end{tabularx}
\end{table*}

\tabref{tab:families-full} completes the full comparison on Wan2.2 and CogVideoX.
\method has the highest correlation with human contact predictions in both families by a small margin ($0.372$ and $0.351$), while its accuracy stays slightly below that of DPH-RL on Wan2.2 and of the base model on CogVideoX, consistent with the fact that accuracy rewards committing to the likelier outcome.
Single-video metrics of \method stay within the range of the other post-training methods in both families, and DynamicsBoost again attains the highest dynamic degree, as on Cosmos.
On the Kubric split, the inverse-frequency weight lowers the KS distance by $0.012$ on Wan2.2, $0.003$ on CogVideoX and $0.014$ at 14B, less than on PisaBench because the split is less ambiguous, and EBFT, whose feature repulsion also acts on appearance, has the second-highest dynamic degree.
EBFT is behind DPH-RL on the Kubric split in all three families although it is ahead on PisaBench in two. Its repulsion acts on features of the whole video and also spreads outcomes that the first frame determines, which costs most on the less ambiguous split, as uniform noise over-widens the narrowest PisaBench frames (\appref{app:more}).
On the held-out benchmarks (\tabref{tab:families-full} and \tabref{tab:14b-full}), no difference between \method and the strongest compared method excludes zero in any model, while single comparison is worse than the base model on CaliBench in Wan2.2 and at 14B, as on Cosmos-2B. CogVideoX is the exception, where single comparison slightly improves on its less concentrated base model.

\begin{table}[t]
  \centering
  \caption{\textbf{Held-out and single-video metrics for Cosmos-Predict2.5-14B.}
  Complements \tabref{tab:scope} under the same protocol, with the metrics of \tabref{tab:families-full}.}
  \label{tab:14b-full}
  \tablestyle
  \begin{tabularx}{\linewidth}{l||YY|Y|Y|Y|Y|Yc}
    \thickhline
    \multirow{2}{*}{\hd{Method}} & \multicolumn{2}{c|}{\hd{Physion}} & \hd{Kub.$^\ddagger$} & \hd{Pisa.} & \hd{Ph.-IQ} & \hd{VB.} & \hd{PAW.} & \hd{Cali.} \\
    \cline{2-9}
     & Pear.\up & Acc.\up & KS\dn & L2\dn & Score\up & Dyn.\up & TVD\dn & mnTV\dn \\
    \hline\hline
Base model & 0.371 & 62.0 & 0.384 & 11.2 & 27.9 & 38.6 & 29.4 & 0.592 \\
\cdashline{1-9}
SFT & 0.389 & 62.3 & 0.284 & 8.9 & 28.4 & 39.5 & \textbf{29.0} & 0.598 \\
RLVR-World & 0.372 & 61.5 & 0.326 & 8.3 & 27.9 & 37.1 & 31.7 & 0.619 \\
Flow-GRPO & 0.366 & 61.2 & 0.331 & 8.2 & 28.9 & 37.8 & 32.0 & 0.624 \\
ORO (Seg) & 0.359 & 61.4 & 0.318 & 8.3 & 29.8 & 38.4 & 32.2 & 0.615 \\
ORO (Flow) & 0.368 & 61.0 & 0.325 & 8.1 & 29.4 & 40.3 & 31.1 & 0.620 \\
ORO (Depth) & 0.375 & 61.6 & 0.320 & \textbf{7.9} & 27.9 & 39.9 & 30.7 & 0.626 \\
DynamicsBoost & 0.384 & 62.1 & 0.287 & 8.4 & \textbf{31.2} & \textbf{42.6} & 29.8 & 0.603 \\
DiverseGRPO & 0.381 & 61.7 & 0.321 & 8.4 & 27.4 & 41.0 & 29.1 & 0.596 \\
DPH-RL & 0.385 & \textbf{62.4} & 0.279 & 8.5 & 26.0 & 38.9 & 29.6 & 0.601 \\
\hline\grp{9}{Sibling pairing with a quadratic score}\\
EBFT & 0.380 & 61.8 & 0.297 & 8.7 & 27.7 & 41.4 & 29.5 & 0.599 \\
\method, $M=0$ & 0.389 & 62.3 & 0.276 & 8.5 & 28.6 & 39.2 & 29.4 & 0.596 \\
\hline\hline
\rowcolor{rowblue}\method (ours) & \textbf{0.401} & 62.2 & \textbf{0.262} & 8.4 & 28.6 & 39.6 & 29.1 & \textbf{0.590} \\
    \thickhline
  \end{tabularx}
\end{table}

At 14B parameters, every method is trained on the same mixture for 150 updates with three seeds (\tabref{tab:scope} and \tabref{tab:14b-full}).
The 14B base model starts narrower than the 2B model (IQR ratio $0.59$ against $0.71$), single comparison narrows it further ($0.45$ for Flow-GRPO), and \method covers $51\%$ of the distance from the base IQR ratio to one ($0.80$).
Single-video metrics at 14B follow the 2B pattern: \method stays within the range of the other post-training methods (L2 $8.4$ against $7.9$--$8.9\times10^{-2}$), and DynamicsBoost again attains the highest dynamic degree.

\begin{figure*}[t]
  \centering
  \includegraphics[width=\textwidth]{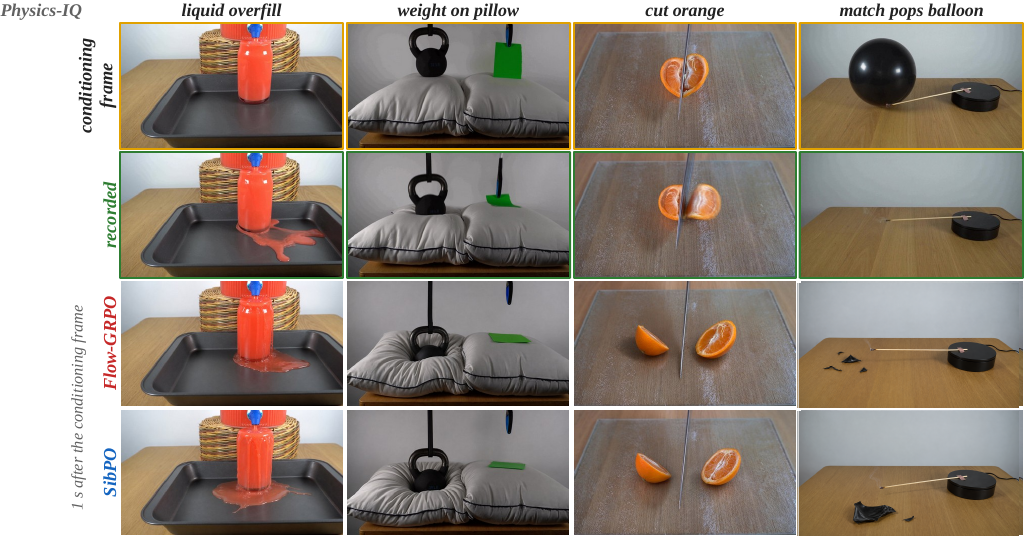}
  \caption{\textbf{Single-video quality on Physics-IQ.}
  Generations one second after the conditioning frame of four real scenes, next to the recorded continuation.
  Flow-GRPO and \method produce physically similar continuations, consistent with their close Physics-IQ scores.}
  \label{fig:physicsiq}
\end{figure*}

\figref{fig:paw} shows two PAWBench scenes whose mechanisms never appear in training.
Flow-GRPO repeats the same pin count or tableware outcome in several of six generations, while \method keeps several outcomes, which matches the held-out columns, where single comparison ends worse than the base model.
\figref{fig:physicsiq} shows that single continuations in real scenes remain physically comparable between Flow-GRPO and \method.

The autoregressive iVideoGPT samples discrete tokens, so siblings share only the key-value cache of the visible history and every token is sampled independently.
Retraining RLVR-World with our seeds reproduces the released checkpoint closely ($0.491$ against $0.486$ MSE and $2.47$ against $2.41$ dispersion).
The Euclidean L2 version of \method keeps $94\%$ of the base dispersion and the L1 plus LPIPS version $90\%$, against $68\%$ for RLVR-World, although the latter gains less in Real2Sim error ($12.2\times10^{-2}$), and the higher MSE of both is the price of not concentrating on the conditional mean, since a model whose generations for one history spread like the data has a larger expected distance to any single recorded future.
Among the added baselines in \tabref{tab:ar}, DPH-RL keeps most of the base dispersion and is the strongest alternative for policy evaluation, while DiverseGRPO keeps less dispersion than DPH-RL because its cluster reward acts on appearance features.
The rank correlation between the absolute Real2Sim error and the centrality $1-2|s-0.5|$ of the real success rate $s$ is $0.38$ for retrained RLVR-World, $0.24$ for DPH-RL and $0.15$ for \method with L2, so the error of single comparison concentrates on policy-task pairs that succeed about half of the time, where a world model that commits each rollout to one outcome estimates a rate of zero or one.

\section{Released Checkpoints}
\label{app:released}

\begin{figure*}[t]
  \centering
  \includegraphics[width=\textwidth]{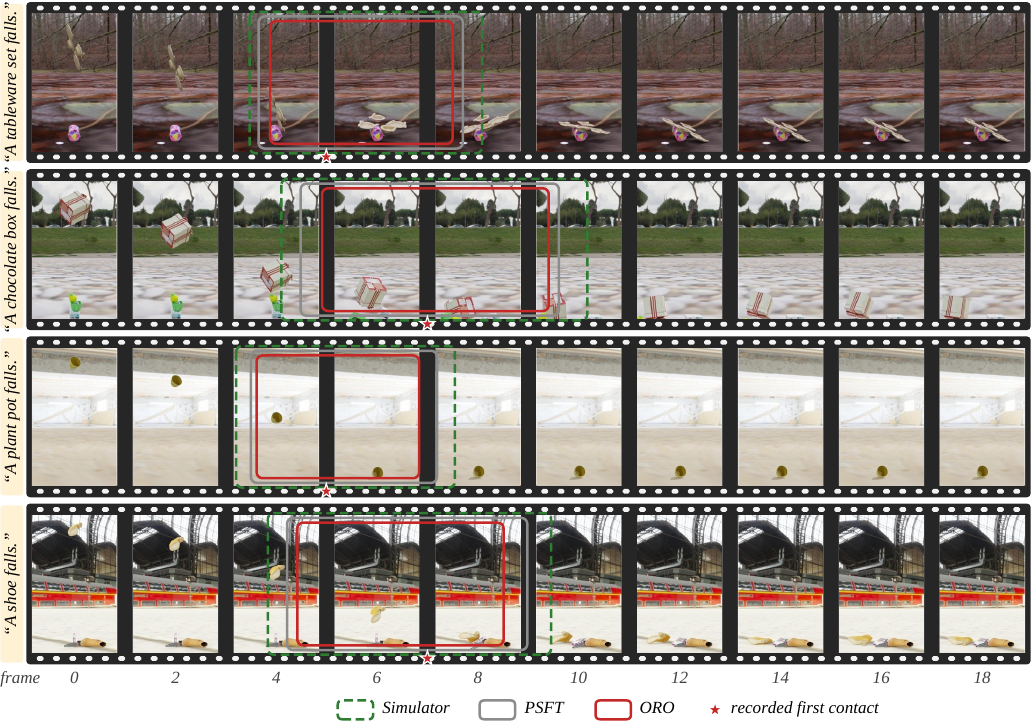}
  \caption{\textbf{Landing-time spread of released PISA checkpoints.}
  Boxes mark the central $80\%$ of first-contact frames over 128 generations for the simulator (green), the supervised checkpoint (gray) and the ORO checkpoint (red) on four first frames.
  The star marks the first contact of the recorded video, shown as a film strip. On every frame the ORO window is the narrowest of the three and contains the recorded contact, while the supervised checkpoint stays closer to the simulated width.}
  \label{fig:released}
\end{figure*}

We evaluate the released Open-Sora~\cite{zheng2024opensora} checkpoints of PISA without retraining, using their published trajectory metrics on 361 real videos and our landing-time measurements from 128 generations for each of the 50 PisaBench frames.
The supervised checkpoint (PSFT) has an L2 error of $7.6\times10^{-2}$, a Chamfer distance (CD) of $0.188$ and an IoU of $0.139$, and its landing times reach a KS distance of $0.277$ with an IQR ratio of $0.81$ and a dispersion ratio $S$ of $0.87$.
The ORO checkpoints with segmentation, flow and depth rewards reach L2 errors of $7.5$, $6.7$ and $6.7\times10^{-2}$, CD of $0.183$, $0.164$ and $0.159$ and IoU of $0.142$, $0.136$ and $0.129$, while their KS distance rises to $0.284$, $0.296$ and $0.291$, their IQR ratio falls to $0.74$, $0.63$ and $0.66$, and $S$ falls to $0.80$, $0.71$ and $0.73$.
The flow and depth variants thus lower L2 and CD most and also narrow most, while the segmentation variant barely changes L2 and narrows least. A reward that is linear in the model distribution predicts this pairing of trajectory gain and concentration.
\figref{fig:released} shows this narrowing on four first frames.
For the ORO checkpoints, the window that contains $80\%$ of generated first contacts is about one frame narrower than the simulated window, which spans two and a half to four frames, and it contains the recorded contact on all four frames, so the model has moved its probability mass toward the frames that the recorded futures of similar frames occupy.

The released RLVR-World checkpoint of iVideoGPT shows the same narrowing with a different architecture and optimizer.
Its single-sample LPIPS is $13.4\times10^{-2}$ against $14.8\times10^{-2}$ for the base model, and its best-of-$N$ LPIPS improves only from $13.4$ to $12.35\times10^{-2}$ between $N=1$ and $N=100$, whereas the base model improves from $14.8$ to $12.05\times10^{-2}$ and overtakes it at $N=50$.
A model whose generations for one history lie close together gains little from more samples, which is the behavior of a narrowed distribution.

\section{Pairing Interventions in Detail}
\label{app:more}

\begin{table}[t]
  \centering
  \caption{\textbf{Interventions on the pairing of \method on Cosmos-Predict2.5-2B.}
  Each row changes one design of \method (gray: leave-one-out baseline, independent noise, $G=8$, with the sibling weight varied in \tabref{tab:family}) and keeps the rest fixed, on trained mechanisms, with three seeds per row.}
  \label{tab:pairing}
  \tablestyle
  \begin{tabularx}{\linewidth}{l||YYY|YYY}
    \thickhline
    \multirow{2}{*}{\hd{Variant}} & \multicolumn{3}{c|}{\hd{PisaBench}} & \multicolumn{3}{c}{\hd{Physion}} \\
    \cline{2-7}
     & KS\dn & IQR\one & $S$\one & Acc.\up & Pear.\up & $\kappa$\up \\
    \hline\hline
\rowcolor{defgray}\method (default) & \textbf{0.241} & \textbf{0.86} & \textbf{0.90} & 61.0 & 0.386 & \textbf{0.198} \\
\hline
\grp{7}{Mean-of-others baseline}\\
$G=4$ & 0.249 & 0.82 & 0.86 & 60.8 & 0.380 & 0.194 \\
$G=8$ & 0.243 & 0.85 & 0.89 & 61.0 & 0.384 & 0.197 \\
$G=16$ & 0.242 & \textbf{0.86} & 0.89 & 61.1 & \textbf{0.387} & 0.197 \\
\hline\grp{7}{Group standard-deviation normalization}\\
$G=4$ & 0.293 & 0.65 & 0.72 & 59.9 & 0.358 & 0.182 \\
$G=8$ & 0.309 & 0.61 & 0.69 & 59.8 & 0.354 & 0.180 \\
$G=16$ & 0.322 & 0.57 & 0.65 & 59.6 & 0.351 & 0.178 \\
Flow-GRPO & 0.329 & 0.55 & 0.62 & 59.6 & 0.349 & 0.178 \\
\hline\grp{7}{Sibling randomness}\\
Shared initial noise & 0.262 & 0.76 & 0.82 & 60.8 & 0.383 & 0.194 \\
Other-history partner & 0.372 & 0.41 & 0.48 & \textbf{61.2} & 0.366 & 0.190 \\
    \thickhline
  \end{tabularx}
\end{table}

\paragraph{Training dynamics.}
The evaluations every 25 updates show how the two pairings reach their endpoints.
Flow-GRPO shrinks the IQR ratio steadily from $0.71$ for the base model to $0.69$, $0.64$, $0.59$, $0.57$ and $0.55$ after 25, 50, 75, 100 and 150 updates, while its KS distance first falls from $0.585$ to $0.301$ at 50 updates and then rises again to $0.329$.
The first phase corrects the landing delay of the base model, which both pairings share, and the second phase is the concentration that single comparison adds once the delay is gone.
\method instead widens the distribution from the start, with IQR ratios of $0.78$, $0.82$, $0.84$, $0.85$ and $0.86$, and its KS distance decreases monotonically to $0.241$.
Early single-video gains are almost identical, with PisaBench simulated-scene IoU of $0.086$ for Flow-GRPO and $0.088$ for \method after 25 updates, so the two methods diverge only after the coarse errors are corrected, at the point where the pairing starts to dominate the update.
The other variants follow the same logic: the cross-history partner narrows from $0.65$ to $0.41$ while its KS distance rises from $0.352$ to $0.372$.
\method and the other sibling-paired variants change their frequency error by less than the variation across seeds over the last 50 updates.
Single comparison is not yet stationary at 150 updates: continued for 450 updates with three seeds, Flow-GRPO narrows further to IQR ratios of $0.53$, $0.51$ and $0.49$ after 200, 300 and 450 updates with a KS distance of $0.342$, while \method reaches $0.87$, $0.87$ and $0.88$ with a KS distance of $0.236$.
The main budget therefore understates the concentration of single comparison, while \method is already stationary.

\figref{fig:advantage} of the main paper shows the advantages of three estimators on one group of the running example and on a PisaBench group.
With six of eight generations showing contact, group standard-deviation normalization gives the no-contact generation the same advantage with and without the sibling term, as Part~3 of \propref{prop:group} states, while the leave-one-out advantage of \method raises it on average.
Part~3 is exact only for two outcome cells. PisaBench readouts have many cells and the Physion readout has a third cell for lost objects, so standardized advantages keep part of the sibling term there, which is consistent with the group standard-deviation rows of \tabref{tab:pairing} lying between \method and Flow-GRPO. Their approach to Flow-GRPO as $G$ grows is measured and has no derivation in this paper.

\begin{figure*}[t]
  \centering
  \includegraphics[width=\textwidth]{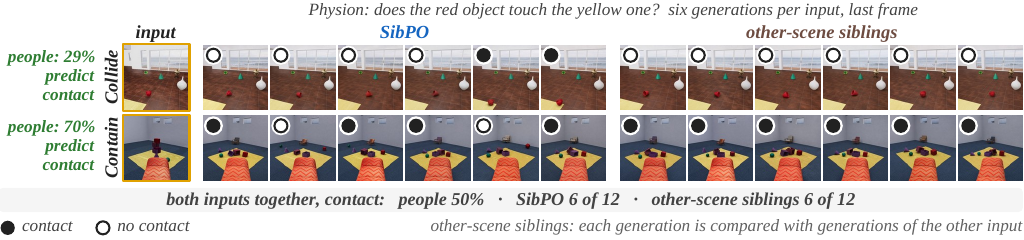}
  \caption{\textbf{Partners from other histories keep the pooled rate and collapse each history.} Two Physion inputs for which $29\%$ and $70\%$ of people predict contact, six generations each.}
  \label{fig:crosshist}
\end{figure*}

The cross-history control of \figref{fig:crosshist} reaches a high Physion accuracy with the lowest IQR ratio, because accuracy rewards committing to the likelier outcome of each history and the pooled frequency is the only quantity that its pairing constrains.
Batch-level diversity terms and distribution-matching objectives that compare sets of generations across histories are subject to the same blind spot.

\begin{figure*}[t]
  \centering
  \includegraphics[width=\textwidth]{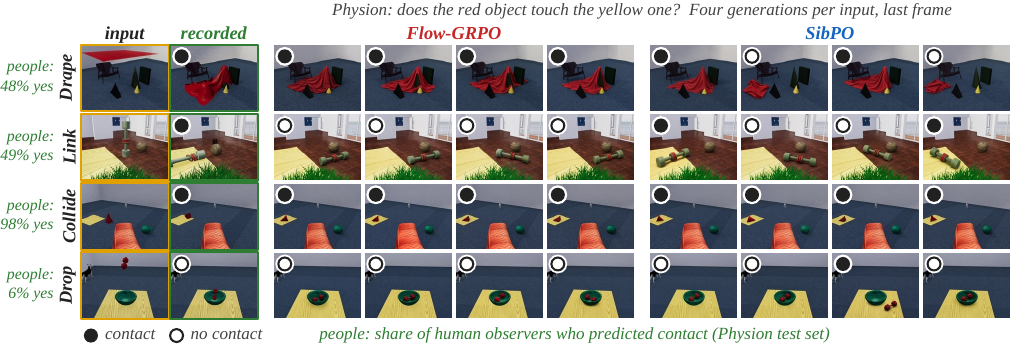}
  \caption{\textbf{Generated contact frequencies follow the ambiguity of the first frame only under \method.}
  Four generations per Physion frame, with the share of people predicting contact on the left.
  Flow-GRPO commits to one outcome even when people are split ($48\%$ and $49\%$), while \method keeps both outcomes there and largely agrees with people on the clear-cut frames.}
  \label{fig:physion}
\end{figure*}

\begin{figure*}[t]
  \centering
  \includegraphics[width=\textwidth]{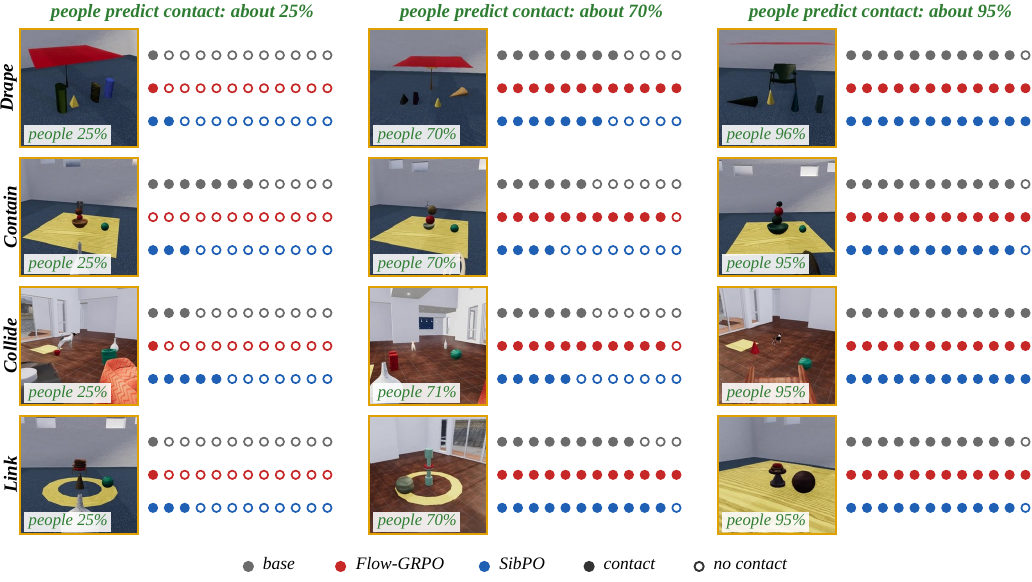}
  \caption{\textbf{Contact frequencies across the range of human agreement.}
  Twelve generations from the base model, Flow-GRPO and \method for Physion frames with about $25\%$, $70\%$ and $95\%$ of people predicting contact, in four scenarios.
  Each dot is the official readout of one generation, filled for contact. Flow-GRPO converts ambiguous frames into near-unanimous outcomes, while the base model and \method keep both outcomes on them.}
  \label{fig:physion-grid}
\end{figure*}

\figref{fig:physion} shows four generations per Physion frame across the range of human agreement, where all models agree with people on clear-cut frames and only \method keeps both outcomes on split ones.
\figref{fig:physion-grid} extends \figref{fig:physion} to twelve frames and twelve generations each.
At $95\%$ human agreement all three models produce contact almost always.
At $25\%$ and $70\%$, Flow-GRPO produces the majority outcome of the recorded futures in at least 11 of 12 generations on every frame.
\method keeps both outcomes on all eight of these frames, but its proportions do not follow the human split frame by frame: at $25\%$ it produces contact in 2 to 5 of 12 generations, and at $70\%$ its contact share ranges from 4 of 12 on Contain to 11 of 12 on Link, so on Contain and Collide most of its generations show no contact although most people predict contact.
Averaged over the eight frames, the absolute gap between the generated and the human contact share is $0.155$ for \method, $0.145$ for the base model and $0.222$ for Flow-GRPO.
On these twelve frames \method therefore keeps the spread of the base model without matching people more closely. Its gains in Brier score and human correlation in \tabref{tab:main} are measured on the full test set and are not visible at this sample size.

\paragraph{Where the gain arises.}
\propref{prop:pairing} predicts that single comparison loses nothing where the true outcome distribution is narrow and loses most where it is wide, because only then does the mode differ from the distribution.
Ordering the 50 PisaBench frames of Cosmos-Predict2.5-2B by simulated landing-time width into five quintiles of ten frames tests this prediction. From the narrowest to the widest quintile, the median IQR ratio is $0.86$, $0.79$, $0.71$, $0.66$ and $0.58$ for the base model, $0.71$, $0.62$, $0.55$, $0.48$ and $0.42$ for Flow-GRPO, $1.36$, $1.12$, $0.95$, $0.81$ and $0.69$ for spread-matched Flow-GRPO, and $0.93$, $0.89$, $0.86$, $0.83$ and $0.79$ for \method.
The per-frame KS advantage of \method over Flow-GRPO grows with the simulated IQR, with Spearman correlations of $0.78$, $0.81$ and $0.62$ on Cosmos-Predict2.5-2B, Wan2.2-TI2V-5B and CogVideoX1.5-5B-I2V.
A gain from better physics or from a uniform change of the falling speed would spread evenly over frames, and a uniform widening at deployment over-widens the narrow frames ($1.36$ for spread-matched Flow-GRPO).
On the held-out transfer benchmarks the scene-level picture follows the intervals: \method has a lower PAWBench TVD than Flow-GRPO on 14 of 20 readable Cosmos scenes, 13 of 21 Wan scenes and 11 of 19 CogVideoX scenes, and a lower CaliBench score on 7, 6 and 4 of 9 scenes.

\paragraph{Uncertainty of the main comparison.}
On trained mechanisms, the PisaBench KS differences between \method and SFT and every method rewarded against the record or preserving diversity have two-level bootstrap intervals below zero on Cosmos-2B, Wan2.2 and 14B.
The closest competitors on Cosmos-2B are DPH-RL, with a difference of $-0.035$ (95\% CI: $-0.052$--$-0.018$), and DynamicsBoost, with $-0.046$ (95\% CI: $-0.064$--$-0.028$), and on CogVideoX the difference against DPH-RL of $-0.009$ (95\% CI: $-0.024$--$0.006$) includes zero.
On the Kubric split, the difference against DPH-RL is $-0.018$ (95\% CI: $-0.031$--$-0.005$) on Cosmos-2B.
The held-out benchmarks have far fewer units, 25 PAWBench scenes and nine CaliBench scenes, so their intervals are wide.
On Cosmos-2B, \method against the base model gives $-1.4$ TVD points (95\% CI: $-4.2$--$1.3$) and $-0.008$ mnTV (95\% CI: $-0.061$--$0.044$), and against DiverseGRPO $-0.1$ TVD points (95\% CI: $-3.4$--$3.2$), while Flow-GRPO against the base model gives $+0.042$ mnTV (95\% CI: $0.003$--$0.083$).
At 14B, the PisaBench KS difference against DPH-RL is $-0.035$ (95\% CI: $-0.056$--$-0.014$), and the held-out differences against the strongest compared method, $+0.1$ TVD points against SFT (95\% CI: $-3.0$--$3.3$) and $-0.006$ mnTV against DiverseGRPO (95\% CI: $-0.041$--$0.029$), include zero, as on Wan2.2 and CogVideoX.
For Physion, the interval of the in-distribution Brier difference against SFT on Cosmos-2B, $-0.011$ (95\% CI: $-0.017$--$-0.005$), resamples test videos and training seeds.
Single-video differences between \method and the single-comparison methods are within the seed variation for L2, IoU and Physics-IQ, so we read them as no measurable change in either direction.

\begin{figure*}[t]
  \centering
  \includegraphics[width=\textwidth]{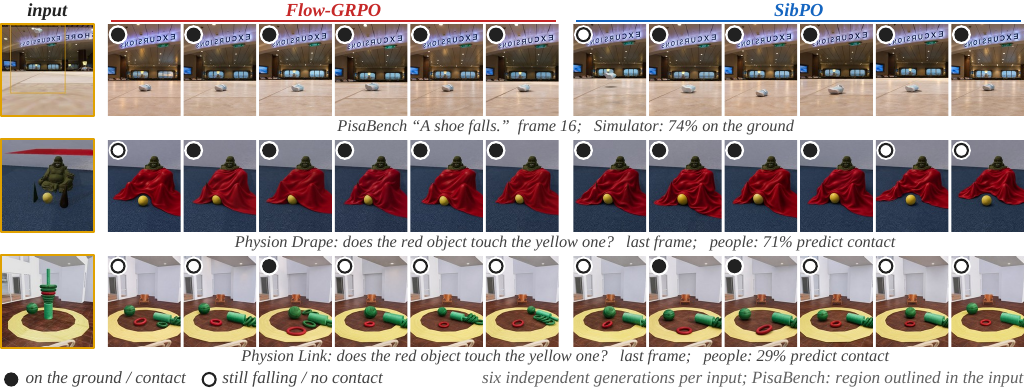}
  \caption{\textbf{Repeated generations of Flow-GRPO and \method on PisaBench and Physion.}
  Six independent generations per input.
  For a PisaBench drop with $74\%$ simulated landings at frame 16 and Physion frames with $71\%$ and $29\%$ human contact predictions, Flow-GRPO produces the majority outcome in five or six of six generations, and \method produces the minority outcome more often, in proportions closer to the reference.}
  \label{fig:qual-pairs}
\end{figure*}

\figref{fig:qual-pairs} complements \figref{fig:qual-all} of the main paper.
In \figref{fig:qual-pairs}, the same pattern holds for a falling object and for two contact scenes with opposite majority outcomes, so the effect does not depend on whether the recorded majority is contact or no contact.
Variation in appearance and variation in the outcome of interest are different properties, and DiverseGRPO, which rewards the former, keeps an IQR ratio close to that of Flow-GRPO ($0.58$ against $0.55$).
The PAWBench coverage metric reflects this difference as well: DiverseGRPO reaches the highest coverage of legal outcomes among the baselines ($67.6\%$ against $64.0\%$ for \method), while its calibration TVD is about the same as that of \method.

\section{Shared Noise and the Measurement of \texorpdfstring{$\Gamma_u$}{Gamma}}
\label{app:gamma}

\begin{table}[t]
  \centering
  \caption{\textbf{Dispersion hidden by shared initial noise.}
  Disagreement rates of landing by frame 16 for replays of the base models with free and tied initial noise, from 16 replay groups on each of the 50 PisaBench frames, at the training sampler noise and at half of it. Share is $\Gamma_u/D_{\mathrm{free}}$.}
  \label{tab:gamma}
  \tablestyle
  \begin{tabularx}{\linewidth}{l||YYYY}
    \thickhline
    \hd{Sampler noise} & $D_{\mathrm{free}}$ & $D_{\mathrm{tied}}$ & $\Gamma_u$ & Share \\
    \hline\hline
\grp{5}{Cosmos-Predict2.5-2B~\cite{nvidia2025cosmospredict25}}\\
Training value & 0.194 & 0.087 & 0.107 & 0.55 \\
Half & 0.188 & 0.063 & 0.125 & 0.66 \\
\hline\grp{5}{Wan2.2-TI2V-5B~\cite{wan2025wan}}\\
Training value & 0.180 & 0.071 & 0.109 & 0.61 \\
Half & 0.181 & 0.052 & 0.129 & 0.71 \\
\hline\grp{5}{CogVideoX1.5-5B-I2V~\cite{yang2025cogvideox}}\\
Training value & 0.209 & 0.121 & 0.088 & 0.42 \\
Half & 0.183 & 0.087 & 0.096 & 0.52 \\
\hline\grp{5}{Cosmos-Predict2.5-14B}\\
Training value & 0.185 & 0.064 & 0.121 & 0.65 \\
Half & 0.170 & 0.047 & 0.123 & 0.72 \\
    \thickhline
  \end{tabularx}
\end{table}

\begin{figure*}[t]
  \centering
  \includegraphics[width=\textwidth]{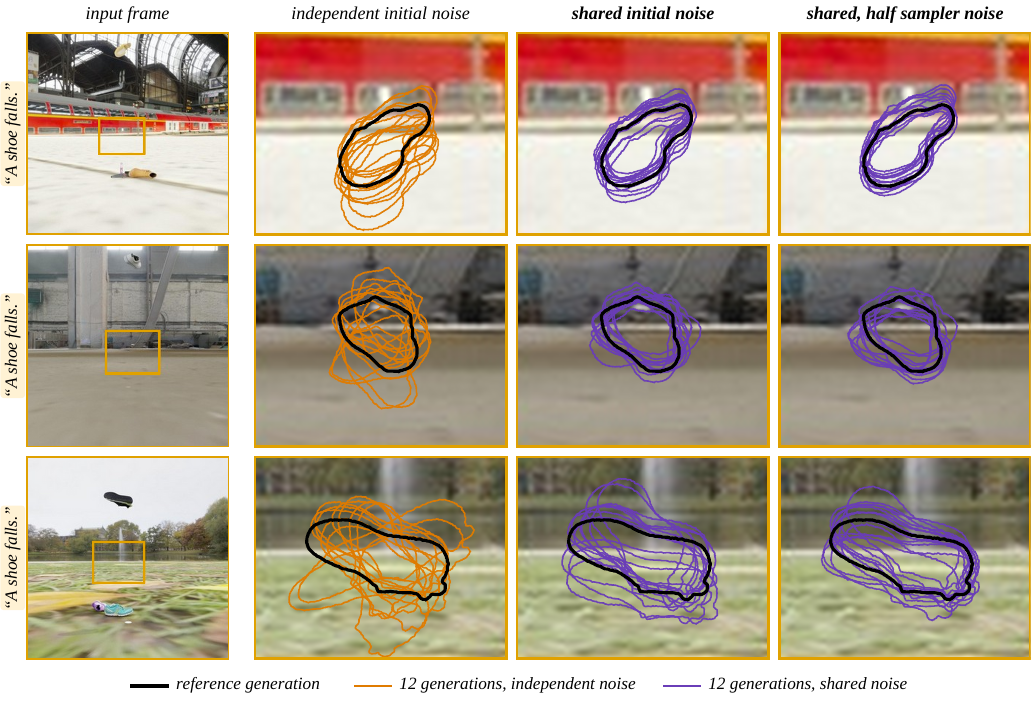}
  \caption{\textbf{Tying the initial noise removes most of the outcome spread.}
  Object outlines at first ground contact for twelve generations from each of three PisaBench frames, compared with a reference generation (black). The input frame on the left marks the enlarged region.
  With independent initial noise (orange) the contacts spread widely, with the initial noise of the reference shared (purple) they stay close to it, and halving the sampler noise tightens them further.}
  \label{fig:shared}
\end{figure*}

$\Gamma_u$ is the part of the model's dispersion that is fixed by the initial noise, an energy-distance analogue of the first-order variance decomposition of Sobol'~\cite{sobol2001global}.
For each frame and replay group, we draw a reference generation, a generation with independent initial noise and a generation that shares the initial noise of the reference with independent later noise, and we record whether each of the two disagrees with the reference about landing by frame 16.
The difference of the two disagreement rates is $\Gamma_u$ for the binary readout (\appref{app:proofs}).
\tabref{tab:gamma} shows that the initial noise determines between $42\%$ and $72\%$ of the outcome disagreement in the four models, with the largest share for the 14B model and the smallest for CogVideoX.
Halving the sampler noise raises the share in every model, since fewer outcomes are left to later noise, although the increase of $\Gamma_u$ itself lies within the bootstrap interval of about $\pm0.040$.
\figref{fig:shared} shows the effect on generations of three frames.

Sharing the initial noise within training groups therefore hides a large part of the sibling signal.
On Cosmos-Predict2.5-2B, the shared-noise variant reaches a KS distance of $0.262$, which is $0.021$ above \method, and on Wan2.2, whose $\Gamma_u$ share is larger, the same variant reaches $0.305$, which is $0.024$ above the $0.281$ of \method, in agreement with the penalty $\Gamma^M_u$ of \eqnref{eq:F2}, which grows with the influence of the initial noise.
The shared-noise variant reaches a plateau of its frequency error within 100 updates, so the training does not remove the dependence on the initial noise in this budget.
With two families, this comparison establishes the direction of the dependence.

\section{Judge Accuracy}
\label{app:judge}

\begin{table}[t]
  \centering
  \caption{\textbf{Judge accuracy sweep on Cosmos-Predict2.5-2B.}
  Training readouts are flipped with probability $\eta$ on generated and recorded videos alike. $\varepsilon$ is the measured readout error on generated videos after flipping.}
  \label{tab:judge}
  \tablestyle
  \begin{tabularx}{\linewidth}{c||Y|YYY|YYY}
    \thickhline
    \multirow{2}{*}{\hd{$\eta$}} & \multirow{2}{*}{\hd{$\bm{\varepsilon}$}} & \multicolumn{3}{c|}{\hd{PisaBench}} & \multicolumn{3}{c}{\hd{Physion}} \\
    \cline{3-8}
     & & KS\dn & IQR\one & $S$\one & Acc.\up & Pear.\up & $\kappa$\up \\
    \hline\hline
\grp{8}{Flow-GRPO (single comparison)}\\
0 & 0.071 & 0.329 & 0.55 & 0.62 & 59.6 & 0.349 & 0.178 \\
0.15 & 0.207 & 0.315 & 0.61 & 0.68 & 60.0 & 0.343 & 0.177 \\
0.30 & 0.339 & 0.308 & 0.66 & 0.73 & 60.3 & 0.357 & 0.183 \\
\hline\grp{8}{\method (sibling pairing)}\\
0 & 0.092 & \textbf{0.241} & \textbf{0.86} & \textbf{0.90} & \textbf{61.0} & \textbf{0.386} & \textbf{0.198} \\
0.15 & 0.209 & 0.259 & 0.82 & 0.87 & 60.4 & 0.375 & 0.192 \\
0.30 & 0.338 & 0.274 & 0.79 & 0.84 & 60.6 & 0.379 & 0.194 \\
    \thickhline
  \end{tabularx}
\end{table}

\begin{figure*}[t]
  \centering
  \includegraphics[width=\textwidth]{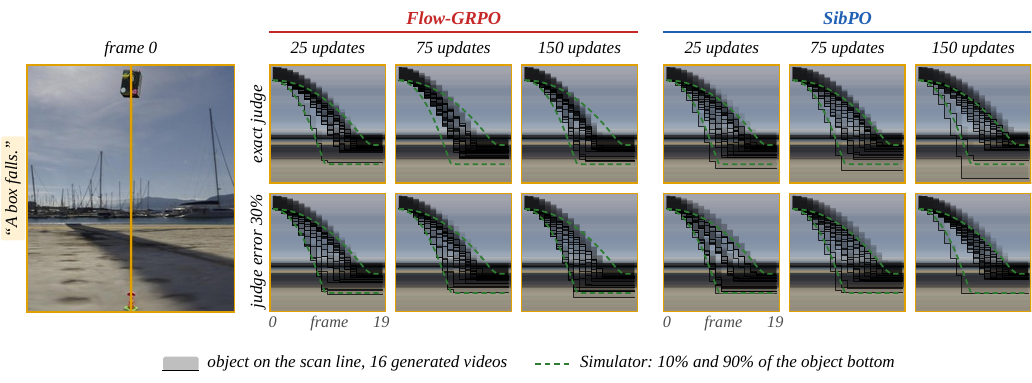}
  \caption{\textbf{A more accurate judge narrows Flow-GRPO and widens \method toward the data.}
  Time slices along a vertical scan line through the falling object, with 16 generations overlaid, after 25, 75 and 150 updates with an exact judge and with $30\%$ judge error. Green lines mark the 10th and 90th percentiles of the simulated trajectory.
  Under the exact judge the trajectories of Flow-GRPO narrow steadily with training, judge error slows this narrowing, and \method covers most of the simulated band under both judges.}
  \label{fig:judge}
\end{figure*}

\tabref{tab:judge} reports the judge sweep.
Flipping with probability $\eta$ raises the measured error on generated and recorded videos alike, so the difference between the two error rates shrinks by a factor $1-2\eta$ while both grow.
The three Flow-GRPO runs order from narrowest to widest as the judge becomes worse, and the three \method runs order the opposite way, and no \method run exceeds the simulated spread.
\figref{fig:judge} shows the same contrast over training: with the exact judge, the overlaid trajectories of Flow-GRPO contract toward the middle of the simulated band as the IQR ratio falls from $0.69$ to $0.55$ between 25 and 150 updates, while with $30\%$ judge error they contract less and end at $0.66$, and \method keeps most trajectories within the simulated band under both judges.

Without flipping, the readout errs on $9.2\%$ of generated videos and on $4.5\%$ of recorded videos, since blur and deformation make generated landings harder to read.
With these rates, the $\beta\to0$ limit of \corref{cor:judge} predicts a shift of $-0.058(1-2q)$, which moves a majority outcome slightly beyond its true probability, that is, a slight narrowing, and this has the sign of the measured IQR deficit of \method, whose ratio stays $0.14$ below one.
The readout asymmetry accounts for only part of this deficit, and the KL penalty and the finite budget account for the rest, a pull toward the narrow reference that the larger weight of \method reduces without removing.

\section{Deployment Sampler and Guidance}
\label{app:deploy}

\begin{table}[t]
  \centering
  \caption{\textbf{Deployment sampler and guidance on Cosmos-Predict2.5-2B.}
  Post-training always uses the stochastic sampler at the default guidance scale $w_0$, whose results are those of \tabref{tab:main}.}
  \label{tab:deploy}
  \tablestyle
  \begin{tabularx}{\linewidth}{l||Y|YYY|YY}
    \thickhline
    \multirow{2}{*}{\hd{Method}} & \multirow{2}{*}{\hd{Setting}} & \multicolumn{3}{c|}{\hd{PisaBench}} & \multicolumn{2}{c}{\hd{Physion}} \\
    \cline{3-7}
     & & KS\dn & IQR\one & $S$\one & Pear.\up & $\kappa$\up \\
    \hline\hline
\grp{7}{Sampler at deployment (guidance $w_0$)}\\
Base model & ODE & 0.597 & 0.70 & 0.77 & 0.359 & 0.182 \\
Flow-GRPO & ODE & 0.351 & 0.54 & 0.61 & 0.350 & 0.179 \\
\method & ODE & 0.272 & 0.79 & 0.84 & 0.381 & 0.195 \\
\hline\grp{7}{Guidance scale (stochastic sampler)}\\
Base model & $w=1$ & 0.651 & 0.82 & \textbf{0.92} & 0.347 & 0.180 \\
Base model & $1.5\,w_0$ & 0.584 & 0.63 & 0.68 & 0.312 & 0.143 \\
Flow-GRPO & $w=1$ & 0.330 & 0.66 & 0.74 & 0.341 & 0.176 \\
Flow-GRPO & $1.5\,w_0$ & 0.338 & 0.49 & 0.56 & 0.352 & 0.181 \\
\method & $w=1$ & 0.283 & \textbf{1.02} & \textbf{1.08} & 0.369 & 0.188 \\
\method & $1.5\,w_0$ & \textbf{0.252} & 0.77 & 0.82 & \textbf{0.383} & \textbf{0.197} \\
    \thickhline
  \end{tabularx}
\end{table}

\tabref{tab:deploy} separates two deployment choices from post-training.
Replacing the stochastic sampler by the deterministic ODE sampler with the same number of steps leaves the base model almost unchanged, as expected from their shared marginals, and it raises the KS distance of \method from $0.241$ to $0.272$, against $0.329$ to $0.351$ for Flow-GRPO.
\method keeps $90\%$ of its advantage under the ODE sampler, so the calibration learned with the stochastic sampler is largely carried by the velocity field, and the stochastic sampler remains the safer choice.

Classifier-free guidance~\cite{ho2022cfg} is a second source of concentration that exists before post-training.
Sampling approximately from $p(x\mid c)^{w}p(x)^{1-w}$ with $w>1$ sharpens the conditional distribution, and the IQR ratio of the base model falls from $0.82$ without guidance to $0.71$ at the default and $0.63$ at $1.5$ times the default.
Wan2.2 and CogVideoX show the same trend, from $0.79$ to $0.64$ and $0.55$ and from $0.89$ to $0.75$ and $0.69$.
\method is calibrated for the guidance scale used during training, so it becomes slightly too wide without guidance (IQR ratio $1.02$) and narrower at the larger scale ($0.77$), while Flow-GRPO is narrow at every scale.
Guidance and pairing therefore act as two independent sources of concentration, and the guidance scale should be kept fixed from post-training to deployment.

\section{Training Diagnostics}
\label{app:monitor}

Two alternative explanations of the main results can be checked on Cosmos-Predict2.5-2B with diagnostics recorded during training.
The first is that \method is better calibrated only because it moves less from the base model.
After 150 updates, \method has a KL of $0.089$ to the reference, below the methods rewarded against the record ($0.091$--$0.159$), but SFT and DPH-RL end lower still ($0.064$ and $0.069$) and remain less calibrated, so the amount of movement does not determine the outcome frequencies.
The second is that the methods exploit the training readout.
On 200 validation videos, the agreement between the training readout and the official judge stays between $0.85$ and $0.87$ for every method after 25 and after 150 updates ($0.856$ and $0.862$ for \method, $0.860$ and $0.857$ for Flow-GRPO), so no method learns to produce videos on which the two programs disagree.
The KL of all reinforcement-learning methods grows fastest during the first 100 updates and flattens afterwards, while the IQR ratio of single comparison keeps decreasing slowly (\appref{app:more}).

\end{document}